\documentclass[10pt]{article} 
\usepackage[accepted]{tmlr}

\usepackage{amsmath,amsfonts,bm}

\def\eqref#1{equation~\ref{#1}}
\def\Eqref#1{Equation~\ref{#1}}

\def\1{\bm{1}}

\DeclareMathAlphabet{\mathsfit}{\encodingdefault}{\sfdefault}{m}{sl}
\SetMathAlphabet{\mathsfit}{bold}{\encodingdefault}{\sfdefault}{bx}{n}

\def\sR{{\mathbb{R}}}

\newcommand{\R}{\mathbb{R}}

\usepackage{hyperref}
\usepackage{url}

\newcommand{\DTW}{\operatorname{DTW}}

\usepackage[utf8]{inputenc}
\usepackage[T1]{fontenc}

\usepackage{microtype}
\usepackage{graphicx}
\usepackage{subfigure}
\usepackage{booktabs} 

\usepackage[ruled]{algorithm2e}

\graphicspath{{../fig/}}

\newcommand\dtwgi{DTW-GI}
\newcommand\stiefel[2]{\mathbb{V}_{#1,#2}}
\def\one{{\mathbf{1}}}
\def\pbf{{\mathbf{p}}}
\def\Pbf{{\mathbf{P}}}
\def\C{{\mathbf{C}}}
\def\U{{\mathbf{U}}}
\def\V{{\mathbf{V}}}
\def\Sigmab{{\boldsymbol{\Sigma}}}

\newcommand{\blk}{\operatorname{blockdiag}}
\newcommand{\diag}{\operatorname{diag}}
\newcommand{\tr}{\operatorname{tr}}

\definecolor{darkjunglegreen}{rgb}{0.3, 0.64, 0.23} 

\newcommand\xbeg{\mathbf{x}_{\rightarrow T'}}
\newcommand\xend{\mathbf{x}_{T' \rightarrow}}
\newcommand\xibeg{\mathbf{x}^{(i)}_{\rightarrow T'}}
\newcommand\xiend{\mathbf{x}^{(i)}_{T' \rightarrow}}
\newcommand\ybeg{\mathbf{y}_{\rightarrow T'}}
\newcommand\yend{\mathbf{y}_{T' \rightarrow}}

\newcommand\yhatend{\hat{\mathbf{y}}_{T' \rightarrow}}

\newcommand{\tv}[1]{#1}

\DeclareMathOperator*{\mingamma}{\min{}^\gamma}

\usepackage{amsthm}
\newtheorem{lemma}{Lemma}[section]

\title{Time Series Alignment with Global Invariances}

\author{\name{Titouan~Vayer}
        \email{titouan.vayer@inria.fr} \\
        \addr{Université Lyon, INRIA, CNRS, ENS de Lyon, UCB Lyon 1, LIP, Lyon, France}
        \AND
        \name{Romain~Tavenard} 
        \email{romain.tavenard@univ-rennes2.fr} \\
        \addr{Université de Rennes, CNRS, LETG, IRISA, Rennes, France}
        \AND
        \name{Laetitia~Chapel}
        \email{laetitia.chapel@irisa.fr} \\
        \addr{Université Bretagne-Sud, CNRS, IRISA, Vannes, France}
        \AND
        \name{Rémi~Flamary} 
        \email{remi.flamary@polytechnique.edu} \\
        \addr{CMAP, Ecole Polytechnique, IP Paris, France}
        \AND
        \name{Nicolas~Courty} 
        \email{nicolas.courty@irisa.fr} \\
        \addr{Université Bretagne-Sud, CNRS, IRISA, Vannes, France}
        \AND
        \name{Yann~Soullard}
        \email{yann.soullard@univ-rennes2.fr} \\
        \addr{Université de Rennes, CNRS, LETG, IRISA, Rennes, France}
        }

\begin{document}
\maketitle

\begin{abstract}
Multivariate time series are ubiquitous objects in signal processing. Measuring a distance or similarity between two such objects is of prime interest in a variety of applications, including machine learning, but can be very difficult as soon as the temporal dynamics and the representation of the time series, {\em i.e.} the nature of the observed quantities, differ from one another. In this work, we propose a novel distance accounting both feature space and temporal variabilities by learning a latent global transformation of the feature space together with a temporal alignment, cast  as a joint optimization problem. The versatility of our framework allows for several variants depending on the invariance class at stake. Among other contributions, we define a differentiable loss for time series and present two algorithms for the computation  of time series barycenters under this new geometry. We illustrate the interest of our approach on both simulated and real world data and show the robustness of our approach compared to state-of-the-art methods.
\end{abstract}


\begin{figure*}[th]
	\centering
	\includegraphics[width=\textwidth]{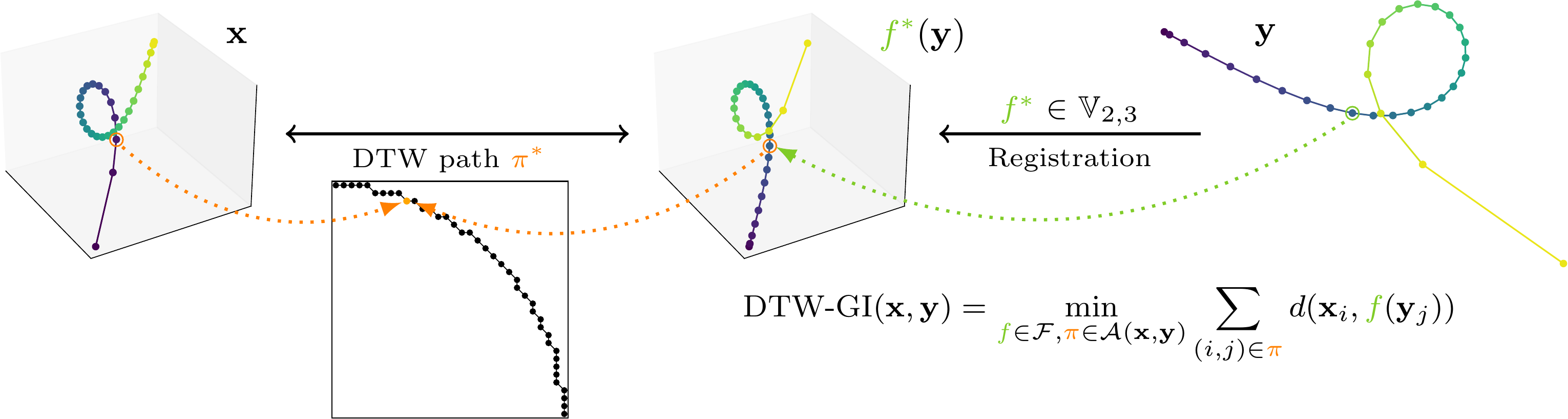}
	\caption{\dtwgi{} aligns time series by optimizing on temporal alignment \tv{$\pi$} (through Dynamic Time Warping) and feature space transformation $f$ \tv{(\textit{e.g.} here in the Stiefel manifold $\mathbb{V}_{2,3}$). In the figure $\pi^{*},f^{*}$ denote the solutions of the optimization problem DTW-GI}. Time series represented here are color-coded trajectories, whose starting (\textit{resp.} end) point is depicted in blue (\textit{resp.} yellow).
		\label{fig:global_scheme}}
\end{figure*}

\section{Introduction}
\label{sec:intro}

Time series are subject to a number of variabilities that make their processing difficult in practice. One of the most well-known example is the temporal shift, usually
handled using the celebrated Dynamic Time Warping (DTW,~\citealt{sakoe1978dynamic}) that aligns, in time, two time series and is invariant to any monotonically increasing temporal map. It has been initially introduced for speech processing applications and is now widely used in a variety of contexts such as human activity recognition \citep{chang2019d3tw}, satellite image analysis \citep{wegner2019dtwsat} or medical applications \citep{huang2020classification,janati2020spatio}.

Another source of variability in time series is 
feature space alterations, that may occur due to a
permutation of sensors, changes in sensor properties or
even different number of sensors.
This problem of heterogeneous
representations, also called distribution
shifts in machine learning, has been studied mostly on non-structured data.
It has been shown that algorithms are notoriously weak \citep{ben2010impossibility}
when it comes to generalizing to out-of-distribution samples, as they rely on the correlations that are found in the training data~\citep{arjovsky2019invariant}. 
Consequently, dedicated paradigms such as domain adaptation \citep{kouw2019review} directly take into account this problem in the learning process. Another approach is to learn with respect to some invariance classes (based on some prior knowledge) in order to be more robust to irrelevant feature transformations \citep{rela_induc_bias,good_inv_2009}. In this work, we aim at tackling this problem in the time series context through the definition of similarity measures that naturally encode desirable invariances. 
More precisely, we introduce similarity measures that are able to deal with both temporal and feature space transformations.

There exist many frameworks to register different spaces under some classes of invariance. In the shape analysis community, matching objects under rigid transformations is a widely studied problem. 
Iterative Closest Point (ICP,~\citealt{CHEN1992145}) is a standard algorithm for such a task.
It acts by alternating two simple steps: (i) matching points using nearest neighbor search and (ii) registering shapes together based on the obtained matches, which is known as the orthogonal Procrustes problem that has a closed form solution~\citep{Goodall:1991}.
This idea is further explored in~\citet{cohen1999earth,alvarez2018towards}, where optimal transport is used to match points in the first step, and a recent extension to objects with a hierarchical structure has been introduced in~\citet{alvarez2019unsupervised} that considers a dedicated invariance class for the registration step.

This heterogeneous setting has also been investigated in the time series context, where the goal is to align series of features lying in different spaces.
One of the most salient track of research in this setting is the Canonical Time Warping (CTW) method.
CTW~\citep{zhou2009canonical} has been introduced for human motion alignment under rigid space transformations.
It consists of temporal alignment (using DTW) of transformed time series, using Canonical Correlation Analysis (CCA) to define the feature space transform.
Few extensions to CTW have been proposed.
GTW~\citep{zhou2012generalized} parametrizes CTW temporal alignments in continuous time instead of relying on DTW.
Deep CTW~\citep{trigeorgis2016deep} extends CTW by learning a feature space embedding (in the form of a neural network) before performing CTW.
Finally, Canonical soft Time Warping applies the CTW methodology to soft alignments (see Section~\ref{sec:dtw} for more details on soft alignments).
In the same vein as CTW, \citet{DENG2020107210} learns an invariant subspace based on DTW alignments.
More recently, GromovDTW~\citep{DBLP:conf/aistats/CohenLTAD21} has been introduced as an extension of the Gromov-Wasserstein distance measure between heterogeneous distributions to the time series context.
GromovDTW relies on time series self-similarities as a way to circumvent the need to compute distances across feature spaces.
Compared to these approaches, our method works by optimizing a map between feature spaces, hence allowing one to (i) add prior information in the form of constraints on the set of allowed maps and (ii) use the computed map for downstream application, as illustrated in our experiments on MoCap data (as described in Section~\ref{sec:xp}).

In more detail, we aim at tackling both temporal and feature space invariances.
To do so, we state the problem as a joint optimization over temporal alignments and feature space transformations, as depicted in Figure~\ref{fig:global_scheme}.
Our general framework allows the use of either DTW or its smoothed counterpart softDTW as an alignment procedure.
Similarly, though rigid transformations of the feature space seem a reasonable invariance class, we show that our method can be used in conjunction with other families of transformations.
Such a framework allows considering the case when time series differ both in length and feature space dimensionality.
We introduce two different optimization procedures that could be used to tackle this problem and show experimentally that they lead to effectively invariant similarity measures.
Our method can also be used to compute meaningful barycenters even when time series at stake do not lie in the same feature space.
Finally, we showcase the versatility of our method and the importance of jointly
learning feature space transformations and temporal alignments on two real-world
applications that are time series forecasting for human motion and cover song identification.


\section{Dynamic Time Warping (DTW)}
\label{sec:dtw}

Dynamic Time Warping (DTW,~\citealt{sakoe1978dynamic}) is an algorithm used to assess similarity between time series, with  extensions to multivariate time series proposed in \citet{ten2007multi, wollmer2009multidimensional}. 
In its standard form, given two multivariate time series  $\mathbf{x} \in \sR^{T_x \times p}$ and $\mathbf{y} \in \R^{T_y \times p}$ of the same dimensionality $p$, DTW is defined as:
\begin{equation}
	\DTW(\mathbf{x}, \mathbf{y}) = \min_{\pi \in \mathcal{A(\mathbf{x}, \mathbf{y})}} \sum_{(i, j) \in \pi} d(\mathbf{x}_i, \mathbf{y}_j)
	\label{eq:DTW}
\end{equation}
where $\mathcal{A(\mathbf{x}, \mathbf{y})}$ is the set of all admissible alignments between $\mathbf{x}$ and $\mathbf{y}$ and $d$ is a ground metric. 
In most cases, $d$ is the squared Euclidean distance, \textit{i.e.} $d(\mathbf{x}_i, \mathbf{y}_j) = \| \mathbf{x}_i - \mathbf{y}_j \|^2$. 

An alignment $\pi$ is a sequence of pairs of time frames which is considered to be admissible iff (i) it matches first (and respectively last) indexes of time series $\mathbf{x}$ and $\mathbf{y}$ together, (ii) it is monotonically increasing and (iii) it is connected (\textit{i.e.} every index from one time series must be matched with at least one index from the other time series).
Efficient computation of the above-defined similarity measure can be performed in quadratic time using dynamic programming, relying on the following recurrence formula:
\begin{equation}
	\DTW(\mathbf{x}_{\rightarrow t_1}, \mathbf{y}_{\rightarrow t_2}) = d(\mathbf{x}_{t_1}, \mathbf{y}_{t_2})
	+ \min 
	\left\{
		\begin{array}{l}
			\DTW(\mathbf{x}_{\rightarrow t_1}, \mathbf{y}_{\rightarrow t_2 - 1}) \\
			\DTW(\mathbf{x}_{\rightarrow t_1 - 1}, \mathbf{y}_{\rightarrow t_2}) \\
			\DTW(\mathbf{x}_{\rightarrow t_1 - 1}, \mathbf{y}_{\rightarrow t_2 - 1})
		\end{array}
	\right.
	\label{eq:dtw:recurrence}
\end{equation}
\tv{where we denote by $\mathbf{x}_{\rightarrow t}$ the time series $\mathbf{x}$ observed up to time $t$}. Many variants of this similarity measure have been introduced.
For example, the set of admissible alignment paths can be restricted to those lying close to the diagonal using the so-called Itakura parallelogram or Sakoe-Chiba band, or a maximum path length can be enforced~\citep{zhang2017dynamic}.
Most notably, a differentiable variant of DTW, coined softDTW, has been introduced in~\citet{cuturi2017soft} and is based on previous works on alignment kernels~\citep{cuturi2007kernel}.
It replaces the min operation in~\Eqref{eq:dtw:recurrence} by a soft-min operator $\min^\gamma$ whose smoothness is controlled by a parameter $\gamma > 0$, resulting in the $\text{DTW}_\gamma$ distance: 
\begin{equation}
	\text{DTW}_{\gamma}(\mathbf{x}, \mathbf{y}) = \mingamma_{\pi \in \mathcal{A(\mathbf{x}, \mathbf{y})}} \sum_{(i, j) \in \pi} d(\mathbf{x}_i, \mathbf{y}_j) = - \gamma \log \left( \sum_{\pi \in \mathcal{A(\mathbf{x}, \mathbf{y})}} e^{-\sum_{(i, j) \in \pi} d(\mathbf{x}_i, \mathbf{y}_j)/\gamma} \right).
	\label{eq:softDTW}
\end{equation}

In the limit case $\gamma=0$, $\min^\gamma$ reduces to a hard min operator and $\text{DTW}_\gamma$ is equivalent to the DTW algorithm.


\section{DTW with Global Invariances}
\label{sec:dtw_gi}

Despite their widespread use, DTW and softDTW are not able to deal with time series of different dimensionalities or to encode feature transformations that may arise between time series. In the following, we introduce a new similarity measure aiming at aligning time series in this complex setting and provide ways to compute associated alignments. We also derive a Fr{\'e}chet mean formulation that allows computing barycenters under this new geometry.

\subsection{Definitions}

Let $\mathbf{x} \in \R^{T_x \times p_x}$ and $\mathbf{y} \in \R^{T_y \times p_y}$ be two time series of length $T_x$ and $T_y$. 
In the following, we assume that the dimension $p_x \geq p_y$. 
In order to allow comparison between time series $\mathbf{x}$ and $\mathbf{y}$, we optimize on a family of functions $\mathcal{F}$ that map $\mathbf{y}$ onto the feature space in which $\mathbf{x}$ lies.
More formally, we define Dynamic Time Warping with Global Invariances (\dtwgi{}) as the solution of the following joint optimization problem: 
\begin{equation}
    \text{DTW-GI}(\mathbf{x}, \mathbf{y}) = \min_{f \in \mathcal{F}, \pi \in \mathcal{A}(\mathbf{x}, \mathbf{y})} \sum_{(i, j) \in \pi} d(\mathbf{x}_i, f(\mathbf{y}_j)),
    \label{eq:dtwgi}
\end{equation}
where $\mathcal{F}$ is a family of functions from $\R^{p_y}$ to $\R^{p_x}$.
Properties (including symmetry) of this similarity measure are detailed in Sec.~\ref{sec:properties}.
Note that this problem can also be written as:
\begin{equation}
    \text{DTW-GI}(\mathbf{x}, \mathbf{y}) = \min_{f \in \mathcal{F}, \pi \in \mathcal{A}(\mathbf{x}, \mathbf{y})} \left\langle \mathbf{W}_\pi , C(\mathbf{x}, f(\mathbf{y})) \right\rangle \label{opt:dtw:dtw_gi}
\end{equation}
where $f(\mathbf{y})$ is a shortcut notation for the transformation $f$ applied to all observations in $\mathbf{y}$,
$\left\langle.,.\right\rangle$ denotes the Frobenius inner
product, $W_\pi$ is defined as:
\begin{equation}
    \forall i \leq T_x, j \leq T_y, \, (\mathbf{W}_\pi)_{i,j} = 
        \begin{cases}
        1 & \text{ if } (i, j) \in \pi\\
        0 & \text{ otherwise}\\
        \end{cases}
\end{equation}
and $C(\mathbf{x}, f(\mathbf{y}))$ is the cross-similarity matrix of squared Euclidean distances between samples from $\mathbf{x}$ and $f(\mathbf{y})$, respectively.
This definition can be extended to the softDTW case of~\Eqref{eq:softDTW} as proposed in the following:
\begin{alignat}{2}
\text{DTW}_\gamma\text{-GI}(\mathbf{x}, \mathbf{y}) =& \min_{f \in \mathcal{F}} \mingamma_{\pi \in \mathcal{A}(\mathbf{x}, \mathbf{y})} \left\langle \mathbf{W}_\pi , C(\mathbf{x}, f(\mathbf{y})) \right\rangle \label{opt:dtw:softdtw_gi} \\
 =&  \nonumber
 \min_{f \in \mathcal{F}} - \gamma \log  \sum_{\pi \in \mathcal{A(\mathbf{x}, \mathbf{y})}} e^{- \left\langle \mathbf{W}_\pi , C(\mathbf{x}, f(\mathbf{y})) \right\rangle/\gamma}. \label{opt:dtw:softdtw_gi_gpos}
\end{alignat}

\begin{figure}[t]
	\centering
	\includegraphics[width=.6\linewidth]{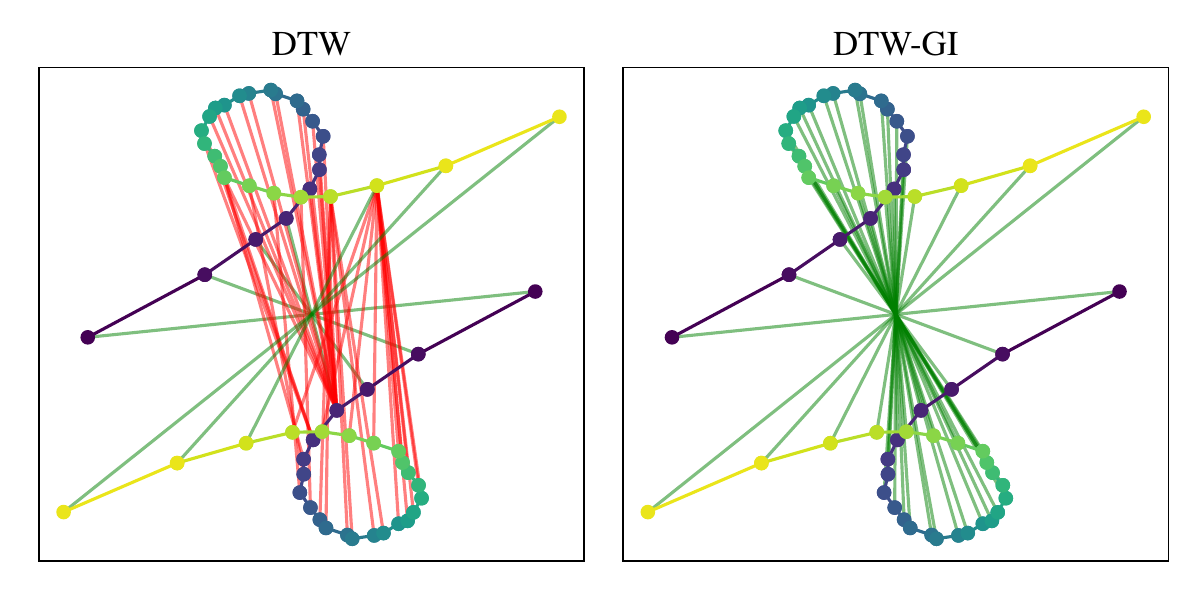}\vspace{-5mm}
	\caption{Example alignments between 2D time series (trajectories in the plane). Color coding corresponds to timestamps.
		Our \dtwgi{} method jointly estimates temporal alignment and global rotation between time series.
		On the contrary, standard DTW alignment fails at capturing feature space distortions and therefore produces a mostly erroneous alignment (matching in red), except at the beginning and end of the time series, whose alignments are preserved thanks to DTW border constraints (see Section~\ref{sec:dtw}).
		\label{fig:alignments}}
\end{figure}

Note that, due to the use of a soft-min operator, \Eqref{opt:dtw:softdtw_gi} is no longer a joint optimization. 

These similarity measures estimate both temporal alignment and feature space
transformation between time series simultaneously, allowing the alignment of time series when
the similarity should be defined up to a global transformation. 
For instance, one
can see in Figure~\ref{fig:alignments} two temporal alignments between two series
in 2D that have been rotated in their feature space. In this case \dtwgi{}, whose
invariant is the space of rotations, recovers the proper alignment whereas DTW fails.

\subsubsection{Properties of \dtwgi{} }
\label{sec:properties}
By definition, \dtwgi{} and soft\dtwgi{} are invariant under any global transformation $T(\cdot)$ such that $\{f \circ T \, | \, f \in \mathcal{F}\} = \mathcal{F}$ (\textit{i.e.} $\mathcal{F}$ is stable under $T$), which motivates the name (soft)DTW with Global Invariances.
Moreover, DTW-GI inherits from some of the classical DTW properties.
First, if for any $f \in \mathcal{F}, f^{-1}$ exists and is in $\mathcal{F}$ (which implies $p_x=p_y$), and if elements of $\mathcal{F}$ are norm preserving operations, then DTW-GI and softDTW-GI are symmetric, since in this case:
\begin{eqnarray}
    \text{DTW-GI}(\mathbf{x}, \mathbf{y})  
        &=&  \min_{f \in \mathcal{F}, \pi \in \mathcal{A}(\mathbf{x}, \mathbf{y})} \sum_{(i, j) \in \pi} d(\mathbf{x}_i, f(\mathbf{y}_j)) \\
        &=& \min_{f \in \mathcal{F}, \pi \in \mathcal{A}(\mathbf{x}, \mathbf{y})} \sum_{(i, j) \in \pi} d(f (f^{-1}(\mathbf{x}_i)), f(\mathbf{y}_j)) \\
        &=& \min_{f \in \mathcal{F}, \pi \in \mathcal{A}(\mathbf{x}, \mathbf{y})} \sum_{(i, j) \in \pi} d(f^{-1}(\mathbf{x}_i), \mathbf{y}_j) \\
        &=& \min_{f \in \mathcal{F}, \pi \in \mathcal{A}(\mathbf{x}, \mathbf{y})} \sum_{(i, j) \in \pi} d(f(\mathbf{x}_i), \mathbf{y}_j) \\
        &=& \min_{f \in \mathcal{F}, \pi \in \mathcal{A}(\mathbf{y}, \mathbf{x})} \sum_{(i, j) \in \pi} d(\mathbf{y}_i, f(\mathbf{x}_j)) = \text{DTW-GI}(\mathbf{y}, \mathbf{x}) .
\end{eqnarray}
Though the constraints on $\mathcal{F}$ for this condition to hold might appear rather strong, this still allows to include standard rigid transformations such as rotations, translations and reflections.

Finally, it is straightforward to see that ${\text{DTW-GI}(\mathbf{x}, \mathbf{x}) = 0}$ for any time series $\mathbf{x}$ as soon as $\mathcal{F}$ contains the identity map.
More generally, regardless of the class of functions $\mathcal{F}$ and for any pair of time series $(\mathbf{x}, \mathbf{y})$, we have ${\text{DTW-GI}(\mathbf{x}, \mathbf{y}) = 0}$ iff there exists $f \in \mathcal{F}$ such that $\mathbf{x}$ and $f(\mathbf{y})$ are equal up to repetitions in the series.


\subsection{Optimization}
\label{sec:optim}

Optimization on the above-defined losses can be performed in several ways, depending of the nature of $\mathcal{F}$.
We now present one optimization scheme for each loss.

\subsubsection{Gradient descent} We first consider the optimization on the softDTW-GI loss (\Eqref{opt:dtw:softdtw_gi}) in the case where $\mathcal{F}$ is a parametric
family of functions, {here denoted $f_\theta$}, that are differentiable with respect to their parameters $\theta$. 
The optimization can be done  with a gradient descent on the
parameters of $f_\theta$.
Since softDTW is smooth (contrary to DTW), this strategy can be used to compute gradients of $\text{DTW}_\gamma\text{-GI}$ \emph{w.r.t.} $\theta$.

Complexity for this approach is driven by (i) that of a softDTW computation and (ii) that of computing $f_\theta(\mathbf{y})$. 
If we denote the latter $c_f$, overall complexity for this approach is hence $O(n_\text{iter} (T_x T_y p_x + c_f))$.
Note that when Riemannian optimization is involved, an extra complexity term has to be added, corresponding to the cost of projecting gradients onto the considered manifold. 
This cost is $O(p_y^3)$ for example when optimization is performed on the Stiefel manifold~\citep{wen2013feasible}, which is an important case for our applications, as discussed in more details in the following.

\subsubsection{Block Coordinate Descent (BCD)}

When DTW-GI (see \Eqref{opt:dtw:dtw_gi}) is concerned, we introduce another strategy that consists in alternating minimization over (i) the temporal alignment and (ii) the feature space transformations. 
We will refer to this strategy as Block Coordinate Descent (BCD) in the following.

Optimization over the alignment path given a fixed transformation $f$ solely consists in a DTW alignment, as described in Section~\ref{sec:dtw}.
For a fixed alignment path $\pi$, the optimization problem then becomes:
\begin{equation}
	\min_{f \in \mathcal{F}} \left\langle \mathbf{W}_\pi , C(\mathbf{x}, f(\mathbf{y})) \right\rangle . \label{eq:optim:fixed_w}
\end{equation}
Recall that $C$ is a matrix of squared distances, which means that the
problem above is a weighted least square problem. 
Depending on $\mathcal{F}$,
there can exist a closed form solution for this problem (\emph{e.g.} when $\mathcal{F}$ is the set of affine maps with no further constraints). 
Let us first note that the matrix $C$
can be rewritten as:
\begin{equation}
    \tv{C(\mathbf{x}, f(\mathbf{y})) = \mathbf{u}_\mathbf{x} \one_{T_y}^{\top} + \one_{T_x} \mathbf{v}_{f,\mathbf{y}}^{\top} - 2 \mathbf{x} f(\mathbf{y})^{\top}}
\end{equation}
where: 
$$ \mathbf{u}_\mathbf{x}=(  \| \mathbf{x}_1 \|^2, \dots, \| \mathbf{x}_{T_x} \|^2 )^\top \text{ and } \tv{{\mathbf{v}_{f,\mathbf{y}}=( \|f(\mathbf{y}_1) \|^2, \dots, \| f(\mathbf{y}_{T_y}) \|^2)^\top} \, \,}$$
\tv{and $\one_{n} = (\underbrace{1, \cdots, 1}_{n \text{ times}})^{\top}$.} 
\tv{In particular, the optimization problem (\ref{eq:optim:fixed_w}) reduces to maximizing $\left\langle \mathbf{W}_\pi , \mathbf{x} f(\mathbf{y})^{\top} \right\rangle$ when $\mathcal{F}$ is a set of norm preserving operations.}


\subsubsection{Estimating $f$ in the Stiefel manifold} Let us consider the special case where
$\mathcal{F}$ is the set of linear maps whose linear operator is an orthonormal
matrix, hence lying on the Stiefel manifold that we denote $\stiefel{p_y}{p_x}$ in the following. \tv{It is defined for $p_y \leq p_x$ as $\stiefel{p_y}{p_x} = \{\mathbf{P} \in \R^{p_x \times p_y}, \mathbf{P}^{\top}\mathbf{P} = \mathbf{I}_{p_y}  \} $ and this} invariance class encodes rigid transformations of the features.
In this case, the optimization problem becomes:
\begin{equation}
    \tv{\min_{\mathbf{P} \in \stiefel{p_y}{p_x}} \left\langle \mathbf{W}_\pi , \mathbf{u}_\mathbf{x} \one_{T_y}^{\top} + \one_{T_x}\mathbf{v}_{\mathbf{P},\mathbf{y}}^\top - 2 \mathbf{x} \mathbf{P}\mathbf{y}^{\top} \right\rangle} \label{opt:dtw:alvarez_Op}
\end{equation}
and we have \tv{$\mathbf{v}_{\mathbf{P}, \mathbf{y}}=( \|\mathbf{P} \mathbf{y}_1 \|^2, \dots, \| \mathbf{P} \mathbf{y}_{T_y} \|^2)^\top = (\|\mathbf{y}_1 \|^2, \dots, \|\mathbf{y}_{T_y} \|^2)^\top = \mathbf{v}_{\mathbf{y}}$} since \tv{for all $j, \|\mathbf{P} \mathbf{y}_j \|^2 = \mathbf{y}_j^{\top} \mathbf{P}^{\top} \mathbf{P} \mathbf{y}_j = \|\mathbf{y}_j\|^2 $} and thus the considered applications are norm-preserving.
Overall, we get the following optimization problem:
\begin{equation}
    \tv{\min_{\mathbf{P} \in \stiefel{p_y}{p_x}} \left\langle \mathbf{W}_\pi , \mathbf{u}_\mathbf{x}\one_{T_y}^{\top} + \one_{T_x}\mathbf{v}_{\mathbf{y}}^\top\right\rangle - 2 \left\langle \mathbf{W}_\pi, \mathbf{x} \mathbf{P}\mathbf{y}^{\top} \right\rangle} \label{opt:decompose:Op}
\end{equation}
\tv{which, since the term $\left\langle \mathbf{W}_\pi ,\mathbf{u}_\mathbf{x}\one_{T_y}^{\top} + \one_{T_x}\mathbf{v}_{\mathbf{y}}^\top\right\rangle$ does not depend on $\mathbf{P}$, is equivalent to solving:}
\begin{equation}
    \tv{\max_{\mathbf{P} \in \stiefel{p_y}{p_x}} \left\langle \mathbf{W}_\pi, \mathbf{x} \mathbf{P}\mathbf{y}^{\top} \right\rangle
    = \max_{\mathbf{P} \in \stiefel{p_y}{p_x}} \left\langle \mathbf{x}^\top \mathbf{W}_\pi \mathbf{y},  \mathbf{P} \right\rangle}
    \label{opt:max_f:alvarez}
\end{equation}
\tv{the last equality being a direct consequence of the cyclic property of the trace}.


\begin{algorithm}
    \caption{\label{algo:bcd}
    Block-Coordinate Descent for \dtwgi{} with Stiefel registration}
    $\mathbf{P} \leftarrow \mathbf{I}_{p_x, p_y}$\;
    \Repeat{convergence}{
        $\mathbf{W}_\pi \leftarrow$ Alignment matrix from $\DTW(\mathbf{x}, \mathbf{y P}^\top)$ \;
        $\mathbf{M} \leftarrow \mathbf{x}^\top \mathbf{W}_\pi \mathbf{y}$ (see \Eqref{opt:max_f:alvarez}) \;
        $\mathbf{U}, \boldsymbol{\Sigma}, \mathbf{V}^\top \leftarrow \operatorname{SVD}(\mathbf{M})$  \;
        $\mathbf{P} \leftarrow \mathbf{U} \mathbf{V}^\top$ \;
    }
\end{algorithm}

As described in~\citet{jaggi2013revisiting}, the latter problem can be solved exactly using Singular Value Decomposition (SVD): 
if $\mathbf{U} \boldsymbol{\Sigma} \mathbf{V}^\top = \mathbf{M}$ is the SVD of a matrix $\mathbf{M}$ of shape $(p_y, p_x)$, then $\mathbf{S}^\star=\mathbf{UV}^\top$ is a solution to the linear problem $\max_{\mathbf{S} \in \stiefel{p_y}{p_x}} \langle \mathbf{S},\mathbf{M} \rangle$.
Note that this method can also tackle the case where $\mathcal{F}$ is an affine map whose linear part lies in the Stiefel manifold by realigning time series means, as discussed for example in~\citet{lawrence2019purely}.
A sketch of the algorithm is presented in Algorithm~\ref{algo:bcd} (for the simplified case where time series means do not have to be realigned).

Interestingly, this optimization strategy where we alternate between time series alignment, \emph{i.e.} time correspondences between both time series, and feature space transform optimization can be seen as a variant of the Iterative Closest Point (ICP) method in image registration~\citep{CHEN1992145}, in which nearest neighbors are replaced by matches resulting from DTW alignment.
Its overall complexity is then \tv{$O(n_\text{iter} (T_x T_y p_x + p_x p_y^2))$}.
This complexity is equal to that of the gradient-descent when $p_x = O(p_y)$. 
However, in practice, the number of iterations required is much lower for this BCD variant, making it a very competitive optimization scheme, as discussed in Section~\ref{sec:xp}.

\paragraph{\tv{Generalizations.}} The algorithms presented above are mainly focused on optimization on the Stiefel manifold. 
Note however that they are not strictly restricted to this case.
Typically, (projected) gradient descent based optimization could be performed \tv{for {softDTW-GI}} on any family of functions parametrized by a neural network.
Regarding the BCD algorithm, it requires a numerically efficient way to compute the optimal feature space transform for a fixed alignment.
In our experiment, we illustrate this in the context of cover song identification, for which aligning song keys is a well-known registration step (see Section~\ref{sec:cover} for details). \tv{Another example is for block-structured linear transformations of the features. More precisely, this corresponds to the case where the features are structured into $k$ groups \textit{i.e.} $p_x = k \cdot q_x$ and $p_y = k \cdot q_y$ with $q_y \leq q_x, k \in \mathbb{N}$ and when one looks for a linear transformation $\mathbf{p} \in \R^{q_x \times q_y}$ that aligns the features of each group. To make the connection with our framework, this coincides with a global transformation $\mathbf{P} \in \R^{p_x \times p_y}$ that can be written as $\mathbf{P} = \blk_k(\pbf)$ where: $$\blk_k(\pbf) = \operatorname{diag}(\underbrace{\pbf, \cdots, \pbf}_{k \text{ times }}) =   \begin{pmatrix}
    \pbf & \mathbf{0} & \dots & \mathbf{0} \\
    \mathbf{0} & \pbf & \dots & \mathbf{0} \\
    \vdots & \vdots & \ddots & \vdots \\
    \mathbf{0} & \mathbf{0} & \dots & \pbf
  \end{pmatrix}.$$
In the special case of $\pbf \in \stiefel{q_y}{q_x}$ it is easy to show that $\Pbf^{\top} \Pbf = \mathbf{I}_{p_y}$ and thus $\Pbf \in \stiefel{p_y}{p_x}$ which also corresponds to a rigid transform of the features but this time structured in blocks. In this situation, the alignment problem becomes:
\begin{equation}
    \min_{\begin{smallmatrix} \mathbf{P} = \blk_k(\pbf) \\ \text{ s.t. } \pbf \in \stiefel{q_y}{q_x} \end{smallmatrix}} \left\langle \mathbf{W}_\pi , \mathbf{u}_\mathbf{x} \one_{T_y}^{\top} + \one_{T_x}\mathbf{v}_{\mathbf{P},\mathbf{y}}^\top - 2 \mathbf{x} \mathbf{P}\mathbf{y}^{\top} \right\rangle. \label{opt:dtw:alvarez_Op_struc}
\end{equation}
As $\Pbf \in \stiefel{p_y}{p_x}$ we can use the same reasoning as above and show that it is equivalent to $\max_{\pbf \in \stiefel{q_y}{q_x}} \left\langle \mathbf{x}^\top \mathbf{W}_\pi \mathbf{y},  \blk_k(\pbf) \right\rangle$. Interestingly the latter problem also admits a closed form expression. More precisely by decomposing $\mathbf{x}^{\top} \mathbf{W}_\pi \mathbf{y}$ into $k$ blocks $\mathbf{C}_{ij}$, each of size $q_x \times q_y$, the solution is given by $\mathbf{S}^{\star} = \mathbf{U} \mathbf{V}^{\top}$ where $\mathbf{U}, \mathbf{V}$ come from the SVD of $\sum_{i=1}^{k} \mathbf{C}_{ii}$, \textit{i.e.} the sum of the diagonal blocks. For more details we refer the reader to the Lemma \ref{lemma:blockdiag_proc} in the supplementary materials. We will illustrate this type of transformation for the alignment of human motion trajectories in Section \ref{sec:ts_forecast}.}

\subsection{Barycenters} 
\label{sec:barycenters}

Let us now assume we are given a set $\{\mathbf{x}^{(i)}\}_i$ of time series of possibly different lengths and dimensionalities.
A barycenter of this set in the \dtwgi{} sense is a solution to the following optimization problem:
\begin{equation}
	\min_{\mathbf{b} \in \R^{T \times p}} \sum_i w_i \min_{f_i \in \mathcal{F}} \DTW(\mathbf{x}^{(i)}, f_i(\mathbf{b})) ,
\end{equation}
where weights $\{w_i\}_i$ as well as barycenter length $T$ and dimensionality $p$ are provided as input to the problem. 
Note that, with this formulation, when $\mathcal{F}$ is the Stiefel manifold, $p$ is supposed to be lower or equal to the dimensionality of any time series in the set $\{\mathbf{x}^{(i)}\}_i$.

In terms of optimization, as for similarity estimation, two schemes can be used.
First, softDTW-GI barycenters can be estimated through gradient descent (and when the set of series to be averaged is large, a stochastic variant relying on minibatches can easily be implemented).
Second, when BCD is used for time series alignment, barycenters can be estimated using a similar approach as DTW Barycenter Averaging (DBA,~\citealt{PETITJEAN2011678}), that would consist in alternating between barycentric coordinate estimation and \dtwgi{} alignments.


\begin{figure*}[t]
	\centering
	\includegraphics[width=.44\linewidth]{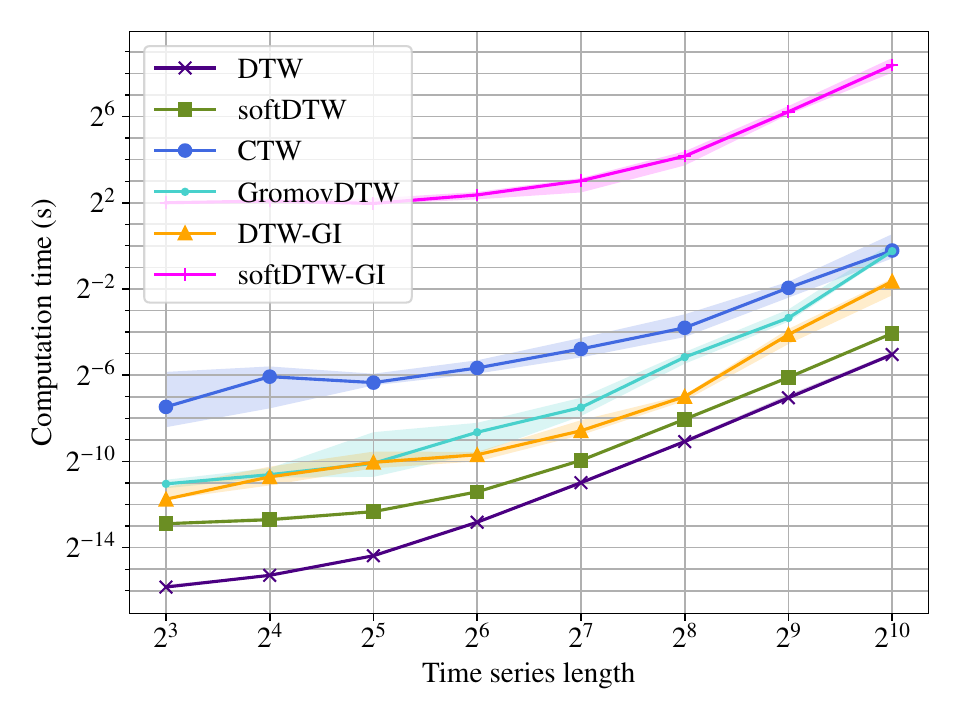} \hspace{1.2cm}
	\includegraphics[width=.44\linewidth]{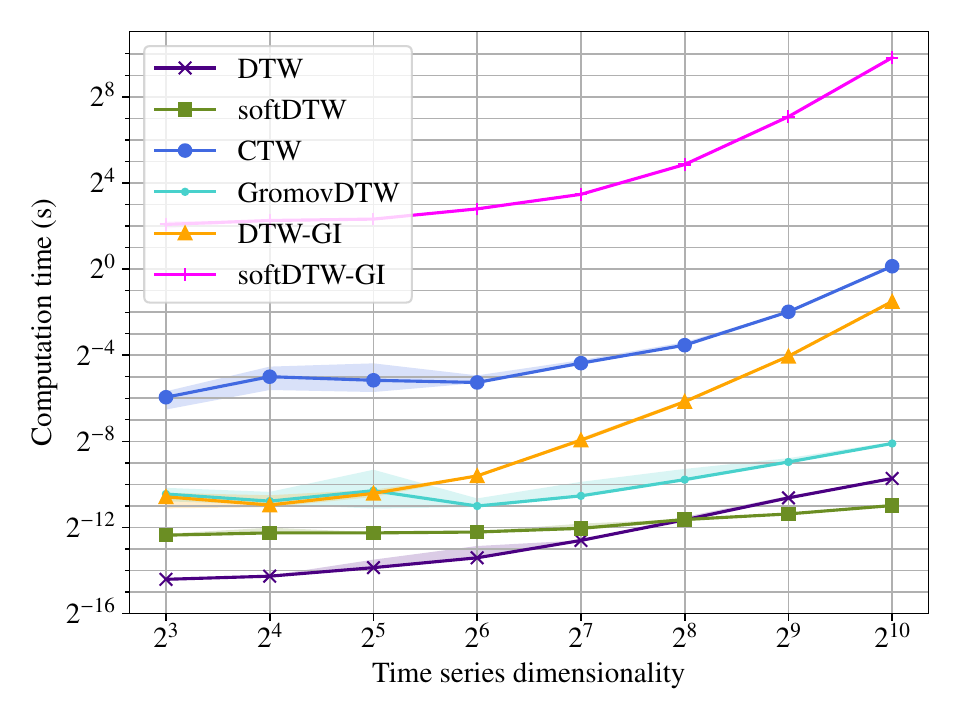}
	\caption{
		Computing time as a function of time series length (left) and dimensionality (right). Solid lines correspond to median values and shaded areas correspond to 20th (\textit{resp.} 80th) percentiles.
		\label{fig:timings}
	}
\end{figure*}

\section{Experiments}
\label{sec:xp}


In this section, we provide an experimental study of \dtwgi{} (and its soft counterpart) on simulated data and real-world datasets.
Unless otherwise specified, the set $\mathcal{F}$ of feature space transforms is the set of affine maps whose linear part lies in the Stiefel manifold.
In all our experiments, tslearn~\citep{tslearn} implementation is used for
baseline methods and gradient descent on the Stiefel manifold is performed using
GeoOpt~\citep{geoopt,becigneul2018riemannian} in conjunction with PyTorch ~\citep{PyTorch2019}.
Open source code of our method will be released upon publication. 

\subsection{Timings}
\label{sec:timings}


We are first interested in a quantitative evaluation of the temporal complexity of our methods.
Note that the theoretical complexity of DTW and softDTW are the same, hence any difference observed in this series of experiments between \dtwgi{} and soft\dtwgi{} would be solely due to their optimization schemes discussed in Section~\ref{sec:optim}. 
In these experiments, the number of iterations for BCD as well as the number of gradient steps for the gradient descent optimizer are set to 5,000.
The BCD algorithm used for \dtwgi{} is stopped as soon as it reaches a
local minimum, while early stopping is used for the gradient-descent variant
with a patience parameter set to 100 iterations. 

We first study the computation time as a function of the length of the time series involved.
To do so, we generate random time series in dimension 8 and vary their lengths from 8 to 1,024 timestamps.

Figure~\ref{fig:timings} (left) shows a clear quadratic trend for all methods presented, except GromovDTW whose complexity is cubic \emph{w.r.t.} the length of the time series due the tensor-matrix multiplication that is involved at each step of its pseudo-Frank-Wolfe algorithm~\citep{DBLP:conf/aistats/CohenLTAD21}. 
Note that \dtwgi{} and its BCD optimizer clearly outperform the gradient descent strategy used for soft\dtwgi{} because the latter requires more iterations before early stopping can be triggered.
Building on this, we now turn our focus on the impact of feature space dimensionality $p$ (with a fixed time series length of 32).
DTW and softDTW baselines are asymptotically linear with respect to $p$.
Similarly, since GromovDTW relies on pre-computed self-similarity matrices, it only linearly depends in the feature space dimensionality for the computation of these self-similarity matrices.
Since feature space registration is performed through optimization on the Stiefel manifold, both our optimization schemes rely on Singular Value Decomposition, which leads to an $O(p^3)$ complexity that can also be observed for both methods in Figure~\ref{fig:timings} (right).
Note also that the CTW baseline is slightly more computationally expensive than \dtwgi{} in practice, even if asymptotic complexities are the same as for \dtwgi{}.

\subsection{Rotational invariance}

\begin{figure*}[t]
	\centering
	\includegraphics[width=.44\linewidth]{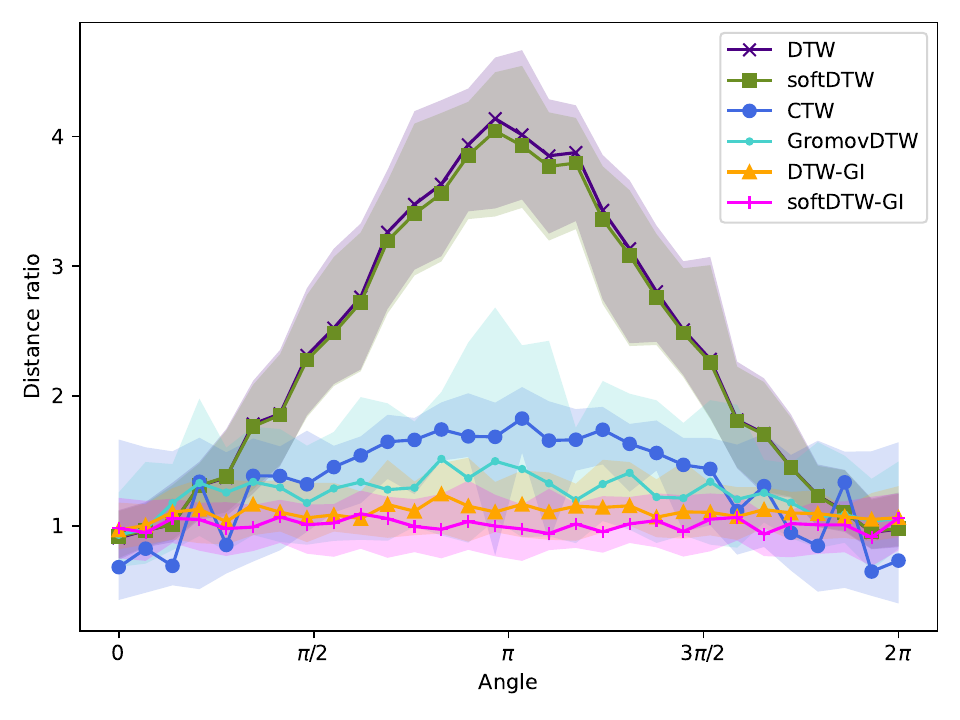}
		\hspace{1.2cm}
	\includegraphics[width=.44\linewidth]{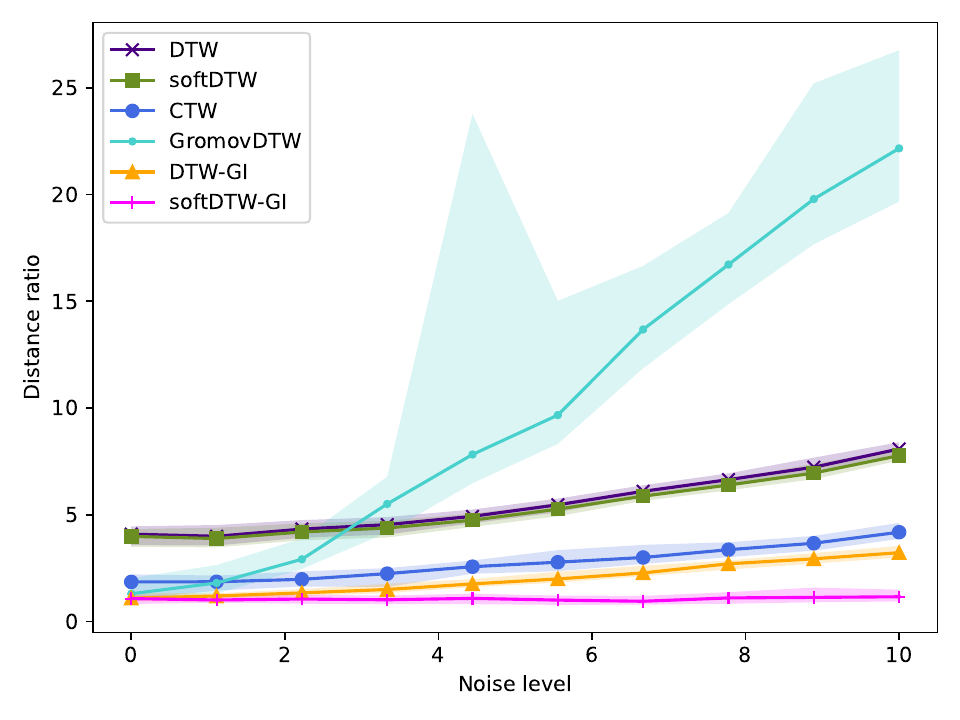}
	
	\caption{
		Illustration of the rotation invariance provided by \dtwgi{}.
		On the left, ratio of the distance to that of a non-rotated pair of spirals is presented as a function of the rotation angle for a fixed noise level.
		On the right, ratio of the distance to that of a non-rotated pair of spirals is presented as a function of the noise level for a fixed rotation angle $\pi$.
		Median distance ratios are reported as solid lines and shaded areas correspond to 20th (\textit{resp.} 80th) percentiles.
		\label{fig:rotation}
	}
\end{figure*}

We now evaluate the ability of our method to recover  invariance to rotation.
To do so, we rely on a synthetic dataset of noisy spiral-like 2D trajectories.
For increasing values of an angle $\alpha$, we generate pairs of spirals rotated
by $\alpha$ with additive Gaussian noise. Alignments between a reference
time series and variants that are subject to an increasing rotation are
computed and repeated 50 times per angle. The ratio of each distance to the distance when $\alpha=0$ is
reported in Figure~\ref{fig:rotation} (left). 
One can clearly see  that the GI counterparts of DTW and softDTW are invariant to rotation in the 2d feature space, while DTW and softDTW are not.
Interestingly, CTW and GromovDTW, that should be invariant to rotation, still exhibit an increase in the loss with the angle $\alpha$, suggesting that their algorithm has more difficulties reaching a global minimum in practice.
Also, when varying the noise level in Figure~\ref{fig:rotation} (right), one
can notice that (soft-)\dtwgi{} are slightly more robust to high levels of noise
than the CTW baseline, while GromovDTW is very sensitive to this noise level
(the GW loss is a quadratic function of its input).

\subsection{Barycenter computation}

\begin{figure*}[t]
	\centering
	\includegraphics[width=\linewidth]{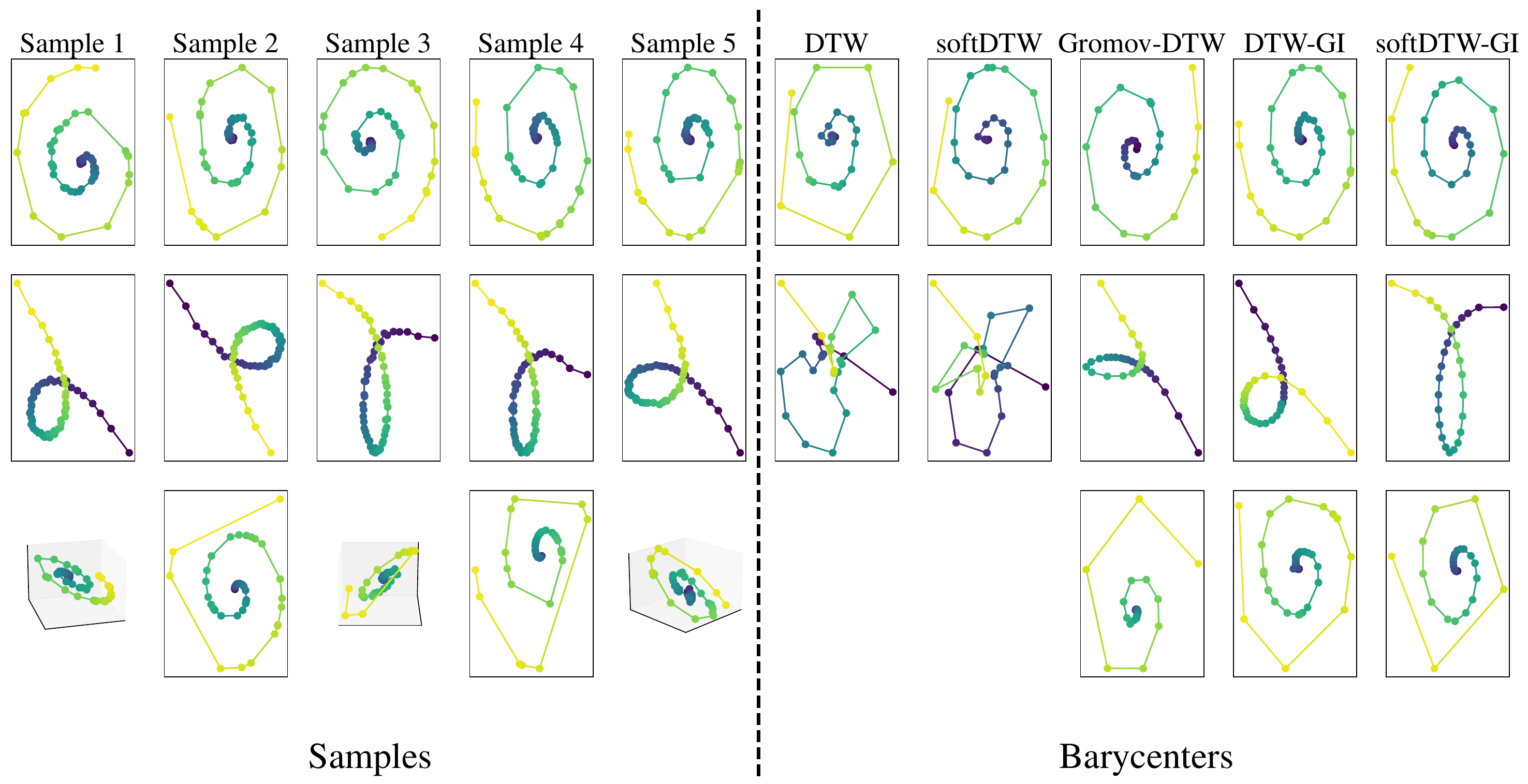}
\vspace{-.5cm}
	\caption{
		Barycenter computation using (i) DTW and softDTW baseline approaches, (ii) the alternative Gromov-DTW~\citep{DBLP:conf/aistats/CohenLTAD21} and (iii) our proposed rotation-invariant \dtwgi{} and soft\dtwgi{}.
		Each row correspond to a different dataset, and the latter one contains both 2D and 3D trajectories, hence cannot be tackled by DTW nor softDTW.
		Trajectories are color-coded from blue (beginning of the series) to yellow (end of the series).
		\label{fig:barycenters}
	}
\end{figure*}

So as to better grasp the notion of similarity captured by our methods, we compute barycenters using the strategy presented in Section~\ref{sec:barycenters}.
Barycenters are computed for 3 different datasets: the first two are
made of 2D trajectories of rotated and noisy spirals or folia, and the third one
is composed of both 2- and 3-dimensional spirals (see samples in
the left part of Figure~\ref{fig:barycenters}).
For each dataset, we provide barycenters obtained by three baseline methods.
DTW Barycenter Averaging (DBA,~\citealt{PETITJEAN2011678}) is used for DTW while softDTW resorts to a gradient-descent scheme to compute the barycenters.
Their GI counterpart use the same algorithms but rely on the
alignments obtained from \dtwgi{} and soft\dtwgi{} respectively. 
Finally, GromovDTW is optimized by alternating between computation of the barycenter self-similarity matrix and alignments, as done in \citet{DBLP:conf/aistats/CohenLTAD21}.
Note that the DTW and softDTW
baselines cannot be used for the third dataset since features of the time series do not lie in the same ambient space.
We would like to emphasize that the barycenter based on GromovDTW only finds a pairwise distance matrix from which the positions of the points must be inferred, for example by applying multidimensional scaling (MDS)~\citep{KruskalWish1978} (as done here and in \citealt{DBLP:conf/aistats/CohenLTAD21}).

For the 2d spiral dataset, all the reconstructed barycenters can be considered as meaningful.
Note however that the outer loop of the spiral (the one that suffers the most from the rotation) is better reconstructed using DTW-GI and softDTW-GI variants.
When it comes to the folia trajectories, that are more impacted by rotations, baseline barycenters fail to capture the inherent structure of the trajectories at stake, while both our methods generate smooth and representative barycenters.
\dtwgi{} and soft\dtwgi{} are even able to recover barycenters when datasets are made of series that do not lie in the same space, as shown in the third row of Figure~\ref{fig:barycenters}.
Finally, in all three settings considered, temporal alignments successfully capture the irregular sampling from the samples to be averaged (denser towards the center of the spiral / loop of the folium).

\begin{figure*}[t]
	\centering
	\includegraphics[width=\linewidth]{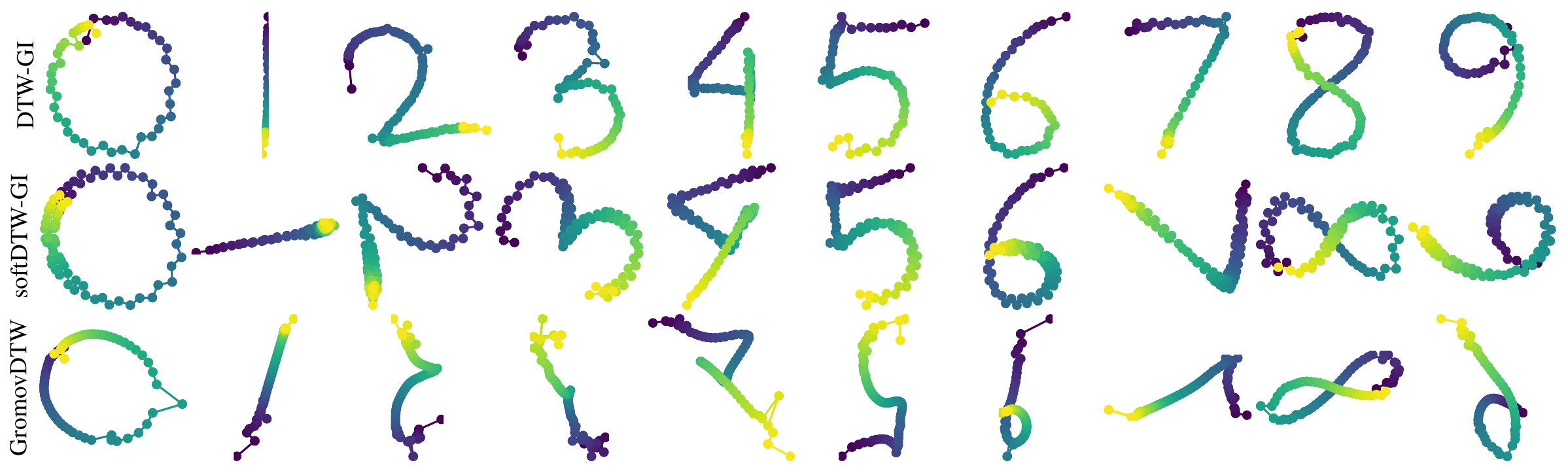}
\vspace{-.5cm}
	\caption{
		Barycenters computed on the RTD Dataset from 50 sample trajectories from each of the 10 classes using DTW-GI, softDTW-GI and GromovDTW.
		Trajectories are color-coded from blue (beginning of the series) to yellow (end of the series).
		\label{fig:barycenters:rtd}
	}
\end{figure*}

The RealSense based Trajectory Digit (RTD) dataset~\cite{s20020376} is made of digit writing trajectories.
In our experiment, we have randomly sampled 50 trajectories per digit class and computed trajectory barycenters using (soft)DTW-GI and GromovDTW.
On such data, expected invariants are rotations and translations.
One can observe that invariance to mirroring strongly impacts GromovDTW's barycenter estimation on classes 4, 7 and 9.
On the other hand, for both DTW-GI and softDTW-GI, we obtain meaningful barycenters that do not suffer from mirroring artifacts and better preserve overall trajectories.

\subsection{Time series forecasting \label{sec:ts_forecast}}

\begin{figure*}
\includegraphics[width=1\linewidth]{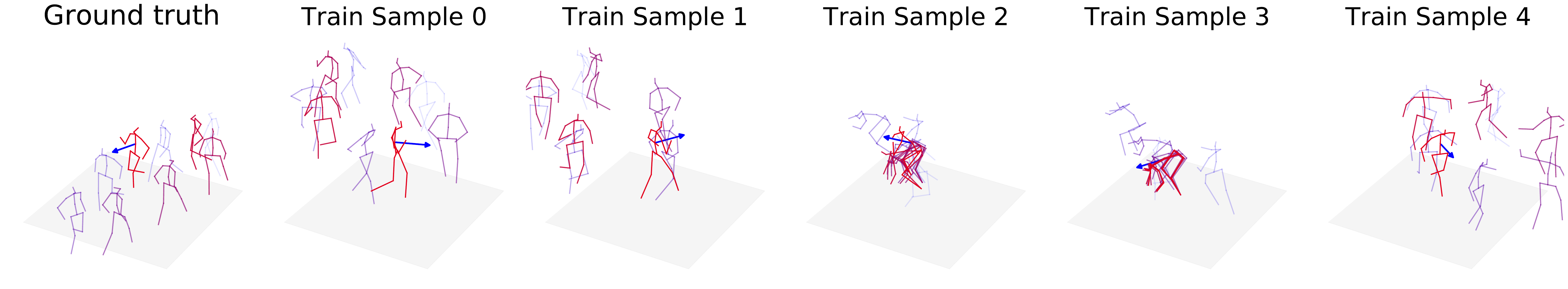}
\hrule
\vspace{6pt}
\includegraphics[width=1\linewidth]{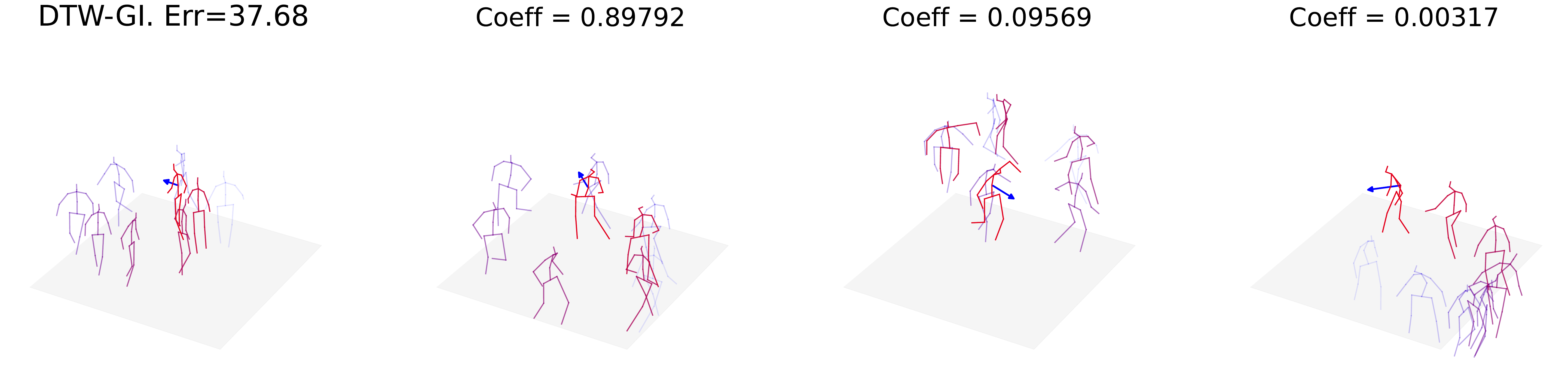}
\hrule
\vspace{6pt}
\includegraphics[width=1\linewidth]{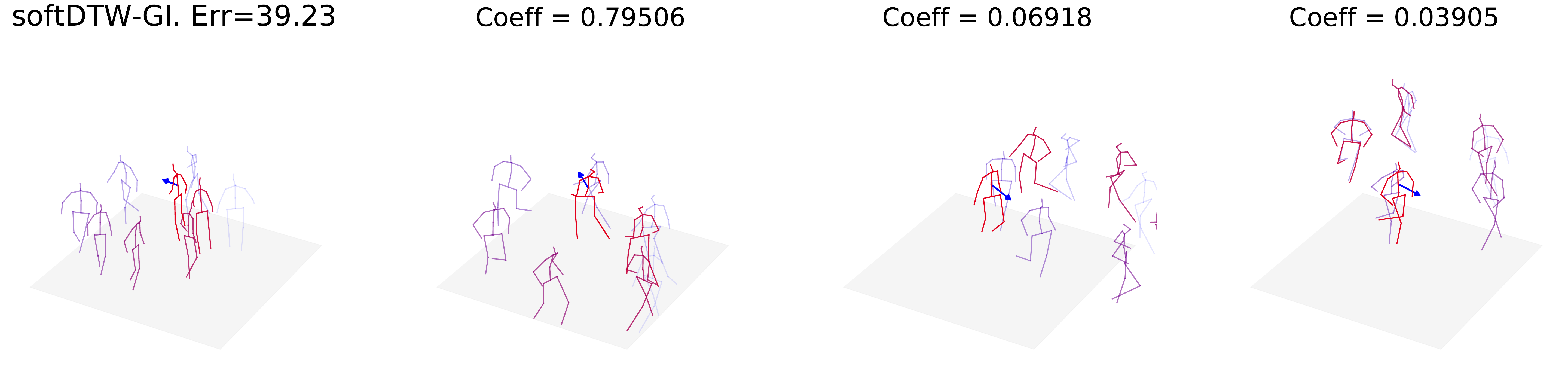}
\hrule
\vspace{6pt}
\includegraphics[width=1\linewidth]{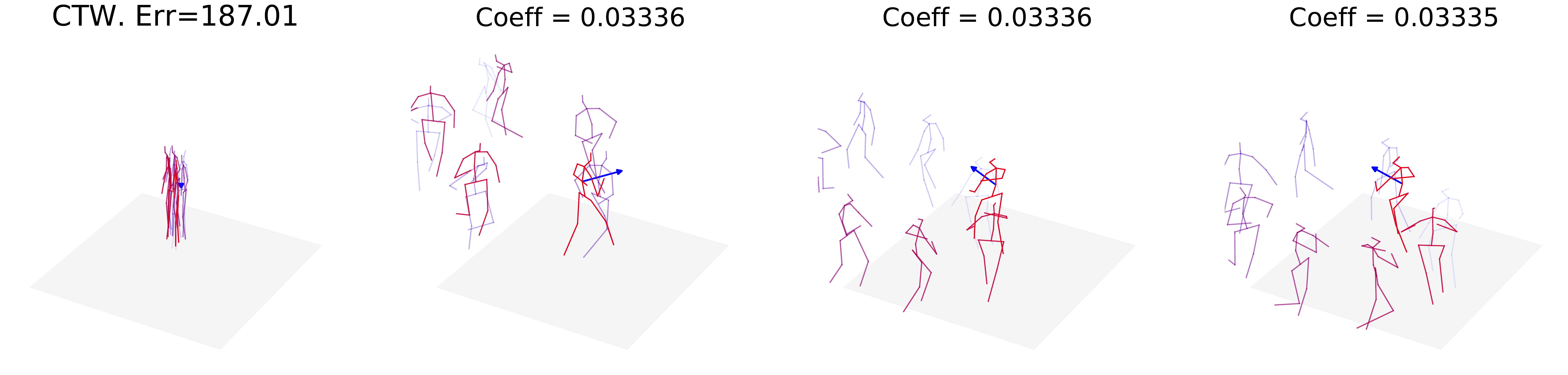}
\hrule
\vspace{6pt}
\includegraphics[width=1\linewidth]{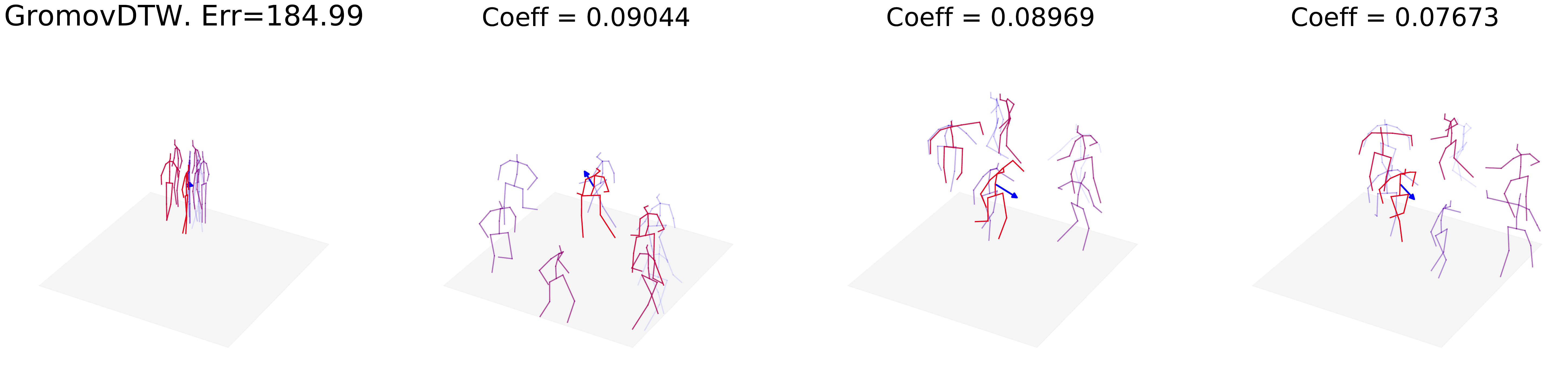}
\caption{Examples of the forecasted subseries. \textbf{(first row)} The first sample is the ground-truth $\yhatend{}$ for the subject $S1$ and then training samples $\xiend{}$ are depicted. \textbf{(from second to last row)} Predictions for the methods {DTW-GI}, {softDTW-GI}, {CTW} and {GromovDTW} (the other methods are given in \ref{fig:mocap_fig2}). In the first column the prediction $\yhatend{}$ is depicted along with the error $\|\yhatend{}-\yend{}\|_2$. In the other columns, we illustrate the first $3$ neighbors \textit{w.r.t.} the method $\xibeg{}$ associated with their coefficients $a_{d}\left(\ybeg{}, \xibeg{}\right)$. For each movement an arrow indicates the orientation of the subject. The beginning of the movement is displayed in shaded blue while the end is displayed in bold red. \label{fig:mocap_fig}}
\end{figure*}

\begin{figure*}
\includegraphics[width=1\linewidth]{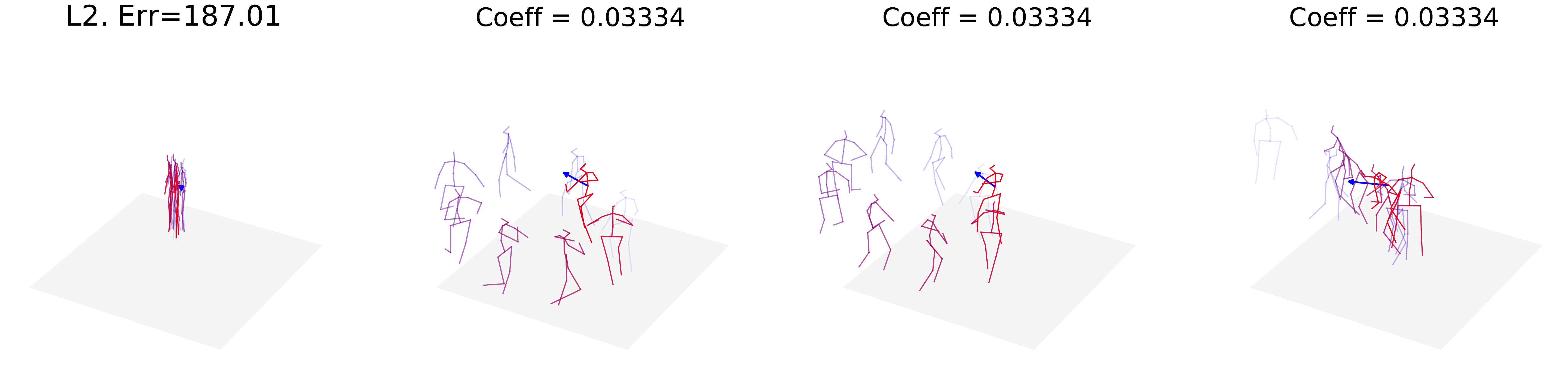}
\hrule
\vspace{6pt}
\includegraphics[width=1\linewidth]{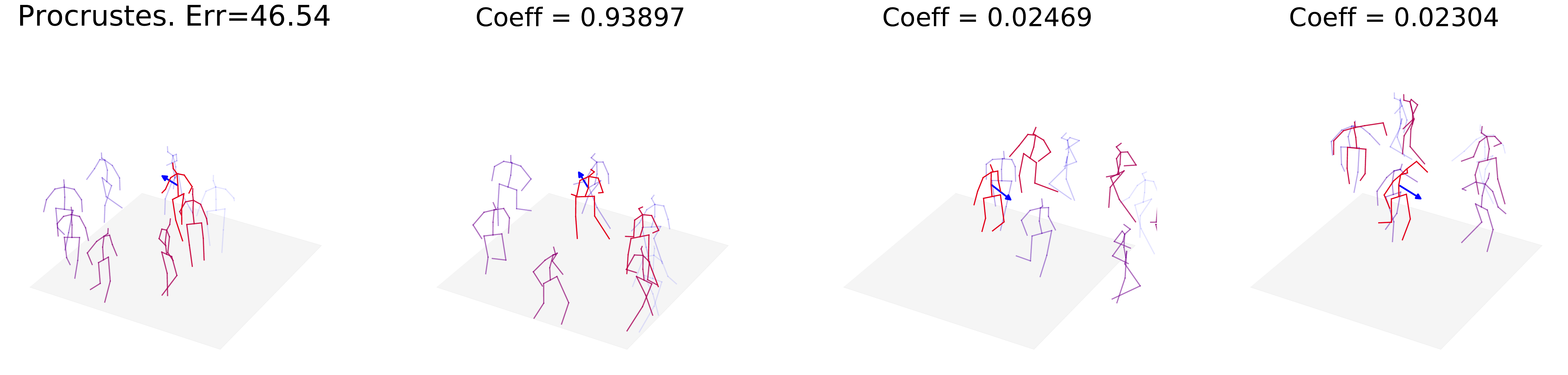}
\hrule
\vspace{6pt}
\includegraphics[width=1\linewidth]{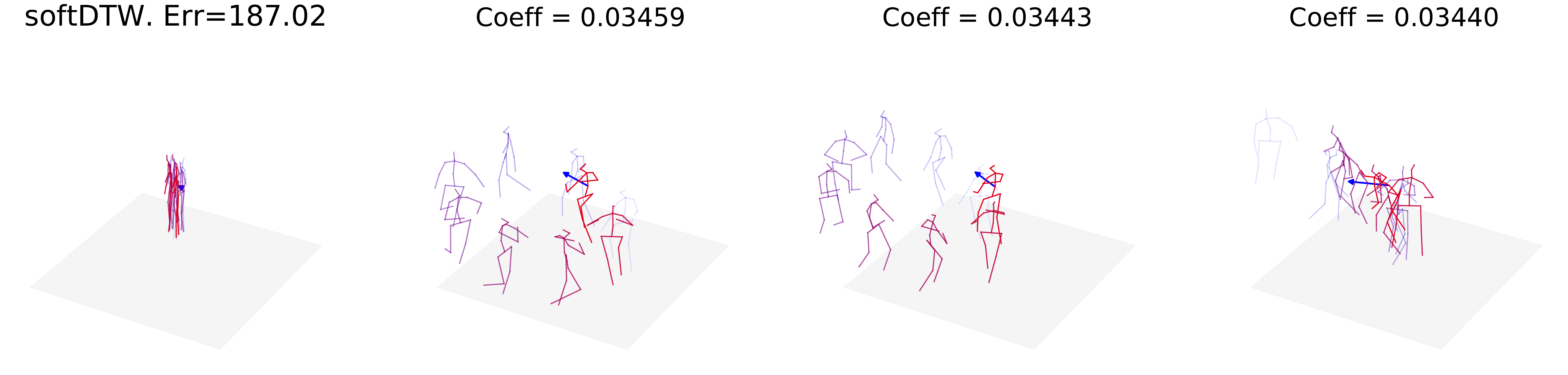}
\hrule
\vspace{6pt}
\includegraphics[width=1\linewidth]{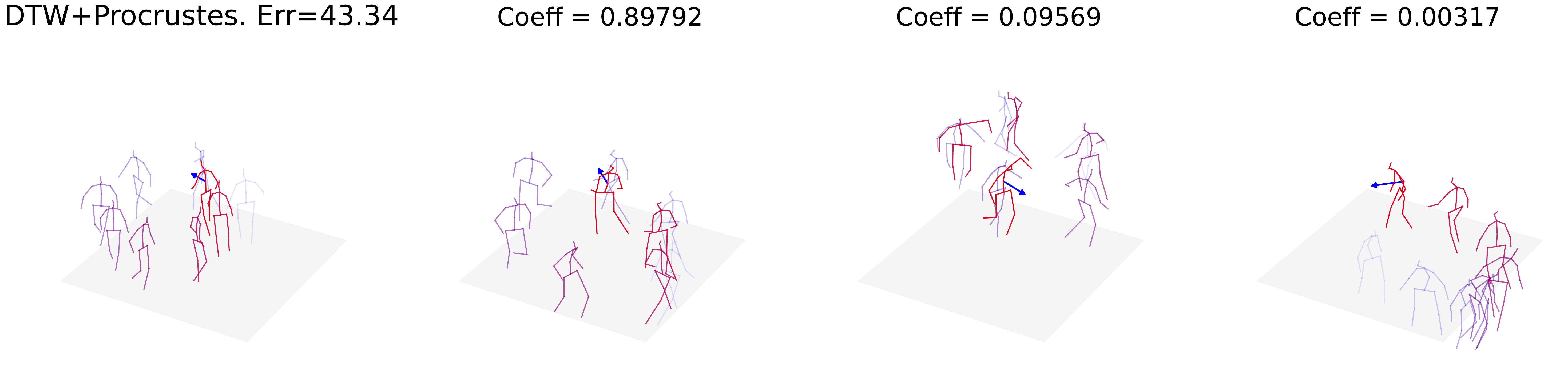}
\hrule
\vspace{6pt}
\includegraphics[width=1\linewidth]{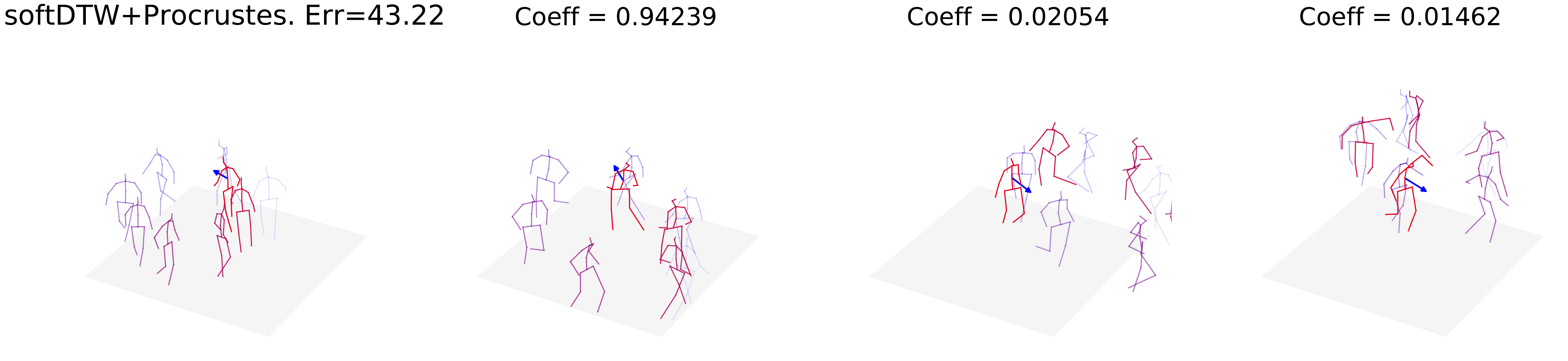}
\caption{Examples of the forecasted subseries for the methods {L2}, {L2+Procrustes}, {softDTW}, {DTW+Procrustes} and {softDTW+Procrustes} \label{fig:mocap_fig2}}
\end{figure*}



To further illustrate the benefit of our approach, we consider a time series forecasting problem \citep{leshape}, where the goal is to infer the future of a partially observed time series. In this setting, we suppose that we have access to a training set of full time series $\mathbf{X}$, with $\mathbf{x}^{(i)} \in \mathbf{X}$ a time series of length $T$ and dimensionality $p_x$, and another test set of partial time series $\mathbf{Y}$ where each $\mathbf{y} \in \mathbf{Y}$ is of length $T' < T$ and dimensionality $p_y$. The goal is to predict the values for timestamps $T'$ to $T$ for each test time series. 
We will denote by $\xbeg$ the beginning of the time series $\mathbf{x}$ (up to time $T'$) and $\xend$ its end (from time $T'$ to time $T$). 

Let $d(\mathbf{y}, \mathbf{x}^{(i)})$ denote a dissimilarity measure between time series $\mathbf{y}$ and $\mathbf{x}^{(i)}$ associated with a transformation $f_{i} \in \mathcal{F}: \mathbb{R}^{p_{x}} \rightarrow \mathbb{R}^{p_{y}}$ that maps the features of $\mathbf{x}^{(i)}$ onto the features of $\mathbf{y}$. This function aims at capturing the desired invariances in the feature space, as described in the previous section.
A typical example is when $d$ is the \tv{$\text{(soft)DTW-GI}$} cost, then the $f_i$ are the Stiefel linear maps which capture the possible rigid transformation between the features. 
We propose to predict the future of a time series $\mathbf{y}$ as follows:
\begin{equation}
    \yhatend{} = \sum_{i} a_{d}\left(\ybeg{}, \xibeg{}\right) f_{i}\left(\xiend{}\right)
    \label{eq:pred}
\end{equation}
where $a_{d}$ is the attention kernel:
\begin{equation}
    a_{d}(\mathbf{y}, \mathbf{x}_i) = \frac{e^{-\lambda d(\mathbf{y}, \mathbf{x}_i)}}{\sum_j e^{-\lambda d(\mathbf{y}, \mathbf{x}_j)}} \label{eq:softmax}
\end{equation}
with $\lambda>0$. 
The prediction is based on the known timestamps for the time series of the
training set and on transformations $f_{i}$ that aim at capturing the latent
transformation between training and test time series. The attention kernel gives
more importance to time series that are close to the time series we want to
forecast \textit{w.r.t.} the notion of dissimilarity $d$. Note that for large
values of $\lambda$, the softmax in~\Eqref{eq:softmax} converges to a
hard max and the proposed approach corresponds to a nearest neighbor imputation.

\subsubsection{Dataset and methodology} 
We use the \textit{Human3.6M} dataset \citep{h36m_pami} which consists of 3.6 million video frames of human movements recorded in a controlled indoor motion capture setting. 
This dataset is composed of 7 actors performing 15 activities (``Walking'', ``Sitting'' ...) twice. 
We are interested in forecasting the 3D positions of the subject joints evolving over time. \tv{More precisely, each data point $\mathbf{x}^{(i)}$ is a time series representing a skeleton of $32$ joints where each joint is described by 3D coordinates. This problem corresponds to $p_x = p_y = 32 \times 3$.} 
We follow the same data partition as~\citet{Coskun_2017_ICCV}: the training set has 5 subjects (S6, S7, S8, S9 and S11) and the remaining 2 subjects (S1 and S5) compose the test set. 
In our experiments, 1) we split the limit frames as follows: we keep the first $T'=300$ timestamps to calculate the coefficient $a_{d}\left(\ybeg{}, \xibeg{}\right)$ and the transformations $f_{i}$ 2) we find the hyperparameter $\lambda$ which gives the best prediction \tv{(\textit{w.r.t.} the $\ell_2$ norm)} for $t \in [T',T_0]$ (where $T_0=400$) 3) the remaining times $[T_0,T']$ are used for the test set. We set the last limit frame as $T'=1100$ which corresponds to predicting $T'-T_0=700$ timestamps, that is predicting $14$ seconds of motion given the initial $8$ seconds.
To emulate possible changes in signal acquisition (\textit{e.g.} rotations of the camera), we randomly rotate the train subjects \textit{w.r.t.} the $z$-axis. We consider the movements of type ``Walking'', ``WalkDog'' and ``WalkTogether'' for the training set and ``Walking'' for the test set. Top row of Figure~\ref{fig:mocap_fig} illustrates samples of movements $\xibeg{}$ resulting from this procedure and the resulting dataset is provided as supplementary material.


\subsubsection{Competing structured prediction methods} 
\tv{We look for global transformations of the features $f_{i}: \mathbb{R}^{32 \times 3} \rightarrow \mathbb{R}^{32 \times 3}$. In order to obtain coherent transformations for each joint of the skeleton, the $f_i$ are structured as $f_i = (\underbrace{g_i, g_i, \cdots, g_i}_{32 \text{ times }}) $ where $g_i : \mathbb{R}^{3} \rightarrow \mathbb{R}^{3}$. In other words, we consider that there is only one global 3D transformation that aligns all the joints of two time series (such as one rotation).} 
We use \tv{DTW-GI and softDTW-GI as our similarity measures and the associated maps $f_{i}$ as described in~\Eqref{opt:dtw:dtw_gi} and in~\Eqref{opt:dtw:softdtw_gi}}. \tv{We compute DTW-GI using the BCD procedure with block-structured rigid transformations of the features as in~\Eqref{opt:dtw:alvarez_Op_struc} ($q_x = q_x = 3, k = 32$ in this context). We calculate softDTW-GI with automatic differentiation as described in Section \ref{sec:optim}. In this experiment we set $\gamma = 0.05$ for the smoothness parameter.}
We compare these methods to \tv{8 baselines}, that correspond to different pairs of time series similarity measure and feature space invariances.
The first two baselines, denoted L2 and softDTW, do not encode any feature space invariance and are based on $\ell_{2}$ and $\text{softDTW}$ similarities respectively.
We also consider a {Procrustes} baseline~\citep{Goodall:1991} defined as: 
\begin{equation}
\label{eq:proc}
d(\ybeg{}, \xibeg{})=\min_{\Pbf, \mathbf{b}} \|(\xibeg{}\mathbf{P}^{\top}+\mathbf{b})-\ybeg{}\|_{2}^{2}
\end{equation}
\tv{where $\Pbf = \blk_{32}(\mathbf{p})$ with $\pbf \in \stiefel{3}{3}$ and $\mathbf{b} = (\overline{\mathbf{b}}, \cdots, \overline{\mathbf{b}})$ with $\overline{\mathbf{b}} \in \mathbb{R}^{3}$.}
The corresponding transformation $f_i$ is the affine map based on the optimal $\mathbf{P}^{\star},\mathbf{b}^{\star}$ found by the previous problem. 
We denote this baseline {L2+Procrustes}. 
\tv{Two other baselines are computed by first registering series using the Procrustes procedure defined above and then using the similarity measure $d(\ybeg{}, \xibeg{})=\DTW(\ybeg{},\xibeg{}\textbf{P}^{\star \top}+\mathbf{b}^{\star})$ and $d(\ybeg{}, \xibeg{})=\DTW_\gamma(\ybeg{},\xibeg{}\textbf{P}^{\star \top}+\mathbf{b}^{\star})$.}
\tv{They are denoted respectively by {DTW+Procrustes} and {softDTW+Procrustes}}. Finally, we also compare with GromovDTW \citep{DBLP:conf/aistats/CohenLTAD21} and CTW \citep{zhou2009canonical}. Note that the methods {L2}, {softDTW}, {GromovDTW} and {CTW} do not provide a transformation of the features of $\xibeg{}$ \textbf{onto} those of $\ybeg{}$ and, as such, we set $f_i=\operatorname{id}$ for all of these methods. 

\subsubsection{Results} 

\begin{table}
	\centering
	\begin{tabular}{c|c}\toprule
		Method & Average test error \\
		\midrule
		{L2} & 183.11 \textpm{} 3.90 \\
		{softDTW}~\citep{cuturi2017soft} & 183.12 \textpm{} 3.90 \\
		\hline
		{CTW}~\citep{zhou2009canonical} & 183.11 \textpm{} 3.90 \\
		{GromovDTW}~\citep{DBLP:conf/aistats/CohenLTAD21} & 181.28 \textpm{} 3.71\\
		\hline
		{L2+Procrustes} & 46.33 \textpm{} 0.21 \\
		{DTW+Procrustes} & 43.31 \textpm{} 0.03  \\
		{softDTW+Procrustes} & 43.90 \textpm{} 0.11 \\
		\hline
		{DTW-GI} (ours) & \textbf{38.05 \textpm{} 0.37} \\
		{softDTW-GI} (ours) & 39.58 \textpm{} 0.34 \\ \bottomrule
		\end{tabular}
		\vspace{3mm}
		\caption{Average error on tests subjects for the time series forecasting on
		the Human3.6M dataset} \label{tab:pred_res}
\end{table}

Qualitative and quantitative results are provided in Figures~\ref{fig:mocap_fig}, \ref{fig:mocap_fig2} and Table~\ref{tab:pred_res} respectively. 
We evaluate, for each test subject, the $\ell_2$ reconstruction loss $\|\yend-\yhatend{}\|_{2}$ between the ground truth time series and its prediction.
Table~\ref{tab:pred_res} displays the average loss on the test subjects based on the best hyperparameter found using the timestamps $[T_0,T']$. 
Figures~\ref{fig:mocap_fig} and \ref{fig:mocap_fig2} present examples of reconstructed movements for the different methods on one test subject as well as the $3$ highest coefficients $a_d$ with the corresponding neighbors.


We observe from the quantitative study that {softDTW}, {L2}, {CTW} and {GromovDTW} lead to the worst reconstruction losses while {L2+Procrustes}, \tv{{DTW+Procrustes}}, {softDTW+Procrustes} and \tv{{DTW-GI}}, {softDTW-GI} lead to the best ones. The results for the first four methods can be explained by the fact that none of them can use an explicit spatial transformation of the feature $f_i$ for the prediction and thus only a simple weighted average of the time-series $\xiend{}$ is realized. This is illustrated for {CTW}, {GromovDTW}, {softDTW} and {L2} in Figures~\ref{fig:mocap_fig} and \ref{fig:mocap_fig2}, where we can see that the prediction tends to shrink. We can also see that \tv{{DTW+Procrustes} and {softDTW+Procrustes}} are superior to a simple {L2+Procrustes} which highlights the importance of temporal realignment. More importantly, the performances of \tv{{DTW-GI} and {softDTW-GI}} are also better than \tv{{DTW+Procrustes},} {softDTW+Procrustes} and {L2+Procrustes} which shows that our \textbf{joint realignment} of time and space has an advantage over a two-step procedure such as {DTW+Procrustes}, {softDTW+Procrustes} which first finds the feature transformation and then aligns series in time. 

Moreover, one can observe qualitatively that {GromovDTW} and {CTW} seem to uniformly average all the different motions to compensate for the lack of reprojection $f_i$. On the contrary, by capturing the possible spatial variability, {L2+Procrustes} and {softDTW+Procrustes} perform reasonably well qualitatively but the predicted movement is slightly less accurate than the one of {softDTW-GI}. This is due to the fact that {L2+Procrustes} or {softDTW+Procrustes} mainly chooses the movement corresponding to the first nearest neighbor ($a_d[1]= 0.94$) while {softDTW-GI} is able to average other dynamics ($a_d[1]= 0.79$). It is somehow a natural conclusion since the optimal transformations found by the {Procrustes} analysis supposes a trivial one-to-one correspondence of the timestamps (\textit{i.e.} $\ybeg(t)$ corresponds to $\xibeg{}(t)$ at \textbf{the same time} $t$) and do not consider the temporal shifts between them. In this way, the method  {L2+Procrustes} leads to unrealistic transformations when the dynamics of movements are not the same. Note that the two-step procedure {softDTW+Procrustes} is only slightly more precise as the feature realignment is independent of the temporal realignment since both are not optimized jointly. On the opposite, {softDTW-GI} method leads to the best qualitative results, highlighting the benefits of our approach over methods that whether discard the temporal variability of the movements ({L2+Procrustes}) or its spatial variability ({softDTW}). 


\subsection{Cover song identification}
\label{sec:cover}

\begin{figure}[t]
	\centering
	\includegraphics[width=.75\linewidth]{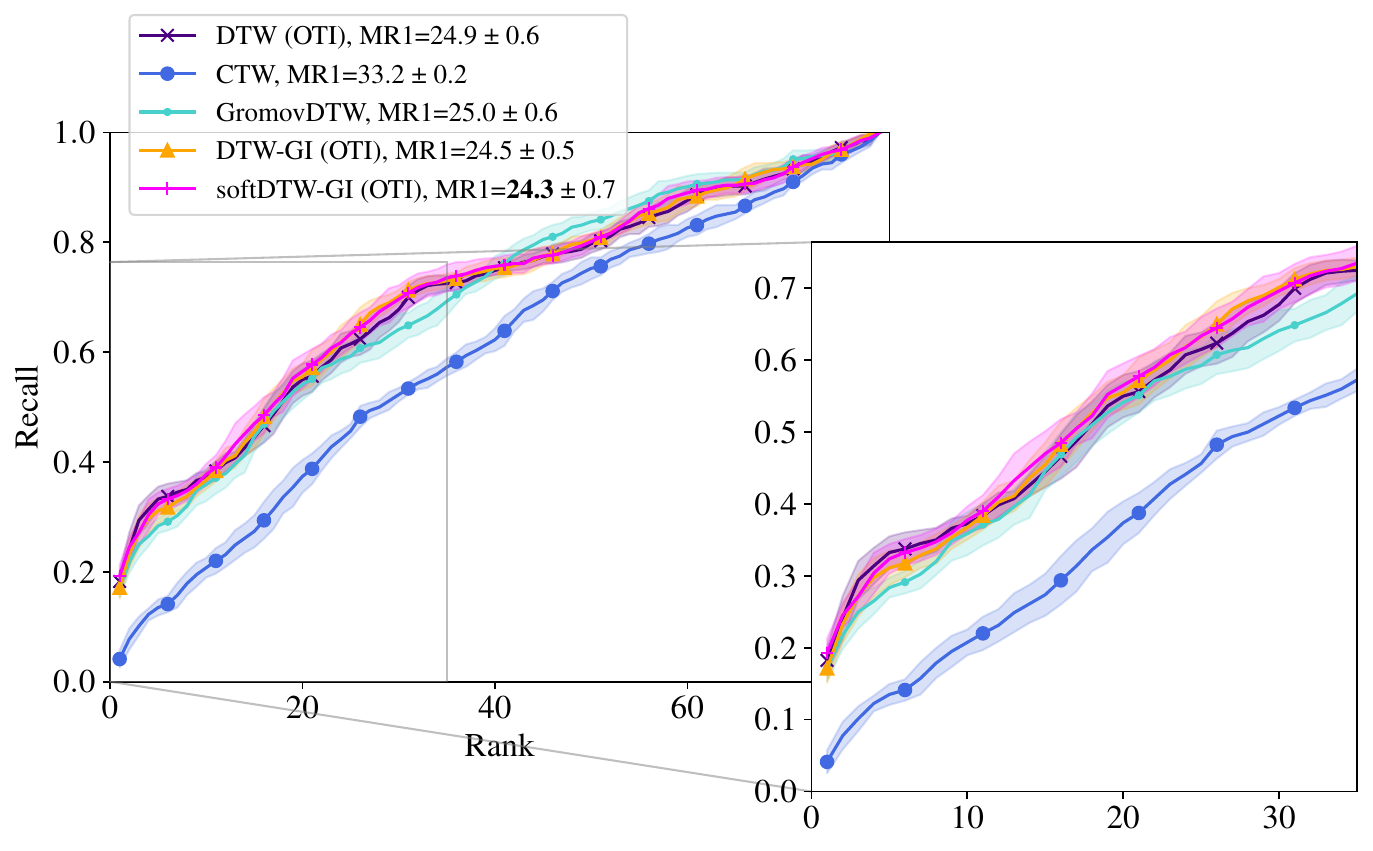}
	\caption{Cover song identification using the covers80 dataset. Methods are compared in terms of recall and results are averaged over 10 train / test set draws. For each method, the shaded area corresponds to one standard deviation around the mean value.
		\label{fig:covers80}
	}
\end{figure}

Cover song identification is the task of retrieving, for a given query song, its covers (\emph{i.e.} different versions of the same song) in a training set.
State-of-the-art methods either rely on anchor matches in the songs and/or on temporal alignments.
In most related works, chroma or harmonic pitch class profile (HPCP) features are usually chosen, as they capture harmonic characteristics of the songs at stake~\citep{heo2017cover}.

For this experiment, we use the covers80 dataset~\citep{ellis20072007} that consists in 80 cover pairs of pop songs and we evaluate the performance in terms of recall.
Since the selection of features is not our main focus, we choose to extract chroma energy normalized statistics (CENS,~\citep{muller2005audio}) over half a second windows.
We compare variants of our method to a baseline that consists in a DTW alignment between songs transposed to the same key using the Optimal Transposition Index (OTI,~\citep{serra2008transposing}).
This OTI computes a transposition based on average energy in each semitone band.
For each competitor, test set songs are ranked based on their distances to query songs.

Figure~\ref{fig:covers80} presents recall scores for compared methods as well as the mean rank of the first correctly identified cover (MR1) that is a standard evaluation metric for the task (used in MIREX\footnote{\url{https://www.music-ir.org/mirex/wiki/2019:Audio_Cover_Song_Identification}} for example). 
\tv{For the sake of readability, performance for \dtwgi{} with feature space optimization on the Stiefel manifold is omitted in this Figure.
In practice, this approach leads to an MR1 of $29.0 \pm 1.0$, which is clearly outperformed by the baseline relying on OTI.
This is because a very common transformation in this setting is when cover songs are played in different keys, which is captured by the OTI transposition strategy.
Similarly, CTW, which does not allow to use prior information about the form of the feature space registration into account, behaves poorly in this setting.}
Interestingly enough, the flexibility of our \dtwgi{} framework allows us to use the OTI strategy.
\tv{Since the registration family $\mathcal{F}$ in this OTI setting is restricted to the set of 12 possible key transpositions, we do not rely on BCD in this case and rather seek for the exact optimum by computing (soft-)DTW for each of the 12 possible transpositions and retain the minimizer.
This way, we are able to compute the optimal transposition index along the alignment path instead of computing it on averaged features, as the ``DTW (OTI)'' baseline does.} 
This leads to a significant improvement of the performance for both \dtwgi{} and its soft counterpart and illustrates both the versatility of our method and the importance of performing joint feature space transformation and temporal alignment.

\tv{Note finally that the performance reached by the competitors in this study is far from what deep models can achieve. ByteCover~\cite{bytecover}, for example, reaches an MR1 of 3.54 on this task, by relying on both a triplet loss and a classification loss derived from a proxy classification task to train a ResNet.
The use of soft\dtwgi{} as an alternative similarity measure in the triplet loss of such an approach is left for future work.}



\section{Conclusion and Perspectives}
\label{sec:conclu}

We propose in this paper a novel similarity measure that can compare time series
across different spaces in order to tackle both temporal and feature space invariances. 
This work extends the well-known Dynamic Time Warping algorithm to deal with time series from different spaces thanks to the introduction of a joint optimization over temporal alignments and space transformations. In addition, we provide a formulation for the computation of the barycenter of a set of times series under our new geometry, which is, to the best of our knowledge, the
first barycenter formulation for a set of heterogeneous time series.
Another important special case of our approach allows for performing temporal
alignment of time series with invariance to rotations in the feature space.

We illustrate our approach on several datasets. First,
we use simulated time series to study the computational complexity of our approach and
illustrate invariance to rotations. Then, we apply our
approach on two real-life datasets for human motion prediction
and cover song identification where invariant similarity measures are shown to improve performance.

Extensions of this work will consider scenarios where features of the series do not lie in a Euclidean space, which would allow covering the case of structured data such as graphs evolving over time, for example. Future works also include the use of our methods in more elaborated models where, following ideas from~\citet{NIPS2019_9338,af722f2cea774ca7a23dc4986804784d},
 softDTW-GI could  be used as a feature extractor in neural networks. It could also serve as a loss to train heterogeneous time series forecasting models~\citep{leshape,cuturi2017soft} or for imitation learning problems as considered in~\citet{DBLP:conf/aistats/CohenLTAD21} where one wants to learn an agent (parametrized by a neural network) that generates trajectories on a different space than the initial ones.

 \subsubsection*{Acknowledgments} 
 This research was supported in part by ANR through the MATS project ANR-18-CE23-0006, by the AllegroAssai project ANR-19-CHIA-0009 and the ACADEMICS grant of the IDEXLYON, project of the Université de Lyon, PIA operated by ANR-16-IDEX-0005. NC is partially funded by the projects OTTOPIA ANR-20-CHIA-0030 AI chair. This research was also supported by 3rd Programme d’Investissements d’Avenir ANR-18-EUR-0006-02. This action benefited from the support of the Chair ‘‘Challenging Technology for Responsible Energy'' led by l’X - Ecole polytechnique and the Fondation de l’Ecole polytechnique. TV gratefully acknowledges the support of the Centre Blaise Pascal’s IT test platform at ENS de Lyon (Lyon, France) for Machine Learning facilities. The platform operates the SIDUS solution \citep{quemener2013sidus}. The results are processed for visualizations using \texttt{matplotlib} \citep{hunter2007matplotlib}. Numerical computations involve \texttt{numpy} \citep{harris2020array}, \texttt{scipy} \citep{virtanen2020scipy} and \texttt{scikit-learn} \citep{scikit-learn} for the CTW implementation.

\bibliography{paper}
\bibliographystyle{tmlr}

\section{Appendix}
\label{sec:appendix}

For a matrix $\pbf$ we note: $$\blk_k(\pbf) = \diag(\underbrace{\pbf, \cdots, \pbf}_{k \text{times}}) =   \begin{pmatrix}
    \pbf & \mathbf{0} & \dots & \mathbf{0} \\
    \mathbf{0} & \pbf & \dots & \mathbf{0} \\
    \vdots & \vdots & \ddots & \vdots \\
    \mathbf{0} & \mathbf{0} & \dots & \pbf
  \end{pmatrix}$$
We will prove the following Lemma:
\begin{lemma}
\label{lemma:blockdiag_proc}
Let $q_x, q_y \in \mathbb{N}$ such that $q_y \leq q_x$ and $k \in \mathbb{N}$. Consider $\C \in \R^{k \cdot q_x \times k \cdot q_y}$. We write the block diagonal decomposition of $\C$ as: $$\C = \begin{pmatrix}
    \C_{11} & \C_{12} & \dots & \C_{1k} \\
    \C_{21} & \C_{22} & \dots & \C_{2k} \\
    \vdots & \vdots & \ddots & \vdots \\
    \C_{k1} & \C_{k2} & \dots & \C_{kk}
  \end{pmatrix}$$ 
where $\C_{ij} \in \R^{q_x \times q_y}$ for $i,j \in \{1, \cdots, k\}$. Consider $\overline{\C} = \sum_{i=1}^{k} \C_{ii} \in \R^{q_x \times q_y}$ the sum of the diagonal blocks of $\C$ and consider the \emph{full} SVD decomposition of $\overline{\C} =  \U \Sigmab \V^{\top}$ where $\U, \Sigmab, \V \in \R^{q_x \times q_y}$. Then the solution to:
\begin{equation}
\label{eq:eq_blk_svd}
\underset{\pbf \in \stiefel{q_y}{q_x}}{\max} \ \langle \C, \blk_k(\pbf) \rangle
\end{equation}
is given by $\pbf^{\star} = \U \V^{\top}$. 
\end{lemma}
\begin{proof}
We have:
\begin{equation*}
\begin{split}
\langle \C, \blk_k(\pbf) \rangle_{F} & = \tr\left(\blk_k(\pbf)^{\top} \C\right) = \tr\left(\begin{pmatrix}
    \pbf^{\top} & \mathbf{0} & \dots & \mathbf{0} \\
    \mathbf{0} & \pbf^{\top} & \dots & \mathbf{0} \\
    \vdots & \vdots & \ddots & \vdots \\
    \mathbf{0} & \mathbf{0} & \dots & \pbf^{\top}
  \end{pmatrix} \begin{pmatrix}
    \C_{11} & \C_{12} & \dots & \C_{1k} \\
    \C_{21} & \C_{22} & \dots & \C_{2k} \\
    \vdots & \vdots & \ddots & \vdots \\
    \C_{k1} & \C_{k2} & \dots & \C_{kk}
  \end{pmatrix}\right) \\
& = \tr\left(\begin{pmatrix}
    \pbf^{\top} \C_{11} & \pbf^{\top}\C_{12} & \dots & \pbf^{\top}\C_{1k} \\
    \pbf^{\top}\C_{21} & \pbf^{\top}\C_{22} & \dots & \pbf^{\top}\C_{2k} \\
    \vdots & \vdots & \ddots & \vdots \\
    \pbf^{\top}\C_{k1} & \pbf^{\top}\C_{k2} & \dots & \pbf^{\top}\C_{kk}
  \end{pmatrix}\right) = \sum_{i=1}^{k} \tr(\pbf^{\top}\C_{ii}) \\
  &= \tr\left( \pbf^{\top}(\sum_{i=1}^{k}\C_{ii}) \right) = \langle \sum_{i=1}^{k}\C_{ii}, \pbf\rangle = \langle \overline{\C}, \pbf\rangle
\end{split}
\end{equation*}
Thus finding the solution to~\Eqref{eq:eq_blk_svd} is the same as finding the solution to $\max_{\pbf \in \stiefel{q_y}{q_x}} \langle \overline{\C}, \pbf\rangle$ which is given by the SVD of $\overline{\C}$ \citep{jaggi2013revisiting}.
\end{proof}

\end{document}